\documentclass[11pt]{article}

\usepackage[preprint]{acl}

\usepackage{times}
\usepackage{latexsym}

\usepackage[T1]{fontenc}

\usepackage[utf8]{inputenc}

\usepackage{microtype}

\usepackage{inconsolata}

\usepackage{graphicx}
\usepackage{enumitem}
\setlist[itemize]{noitemsep, topsep=0pt}
\usepackage{microtype}
\usepackage{graphicx}
\usepackage{booktabs} 
\usepackage{multirow}
\usepackage{arydshln}
\usepackage{amsmath}
\usepackage{amsfonts}
\usepackage[ruled,vlined]{algorithm2e}
\usepackage{subcaption}
\usepackage{setspace}

\title{Enhancing SAE-based Steering via Neighbor Integrated Feature Selection}

\author{
 \textbf{Yutian Liu\textsuperscript{1,2}},
 \textbf{Xu Wang\textsuperscript{1}},
 \textbf{Difan Zou\textsuperscript{1}},
\\
 \textsuperscript{1}The University of Hong Kong,
 \textsuperscript{2}University of Science and Technology of China,
\\
 \texttt{
 liuyutian@mail.ustc.edu.cn, sunny615@connect.hku.hk, dzou@hku.hk}
 }

\newpage
\begin{document}
\maketitle
\begin{abstract}
Sparse autoencoders (SAEs) disentangle model activations into interpretable features and are widely used for steering large language models.
Most existing SAE-based steering methods select features by applying a top-$k$ filter based on statistical scores, assuming that higher-scoring features yield stronger steering effects.
In this paper, we show that this assumption is often invalid, leading to suboptimal feature selection.
Our analysis reveals that effective steering features may be distributed among representationally adjacent, semantically similar groups induced by feature splitting in SAEs.
Within such groups, features may exhibit disparate statistical scores despite having comparable steering influence, causing score-based selection to overlook important features.
Based on these observations, we propose \textsc{Neighbor Integrated Feature Selection} (\textsc{NIFS}), a plug-and-play strategy that leverages representation similarity to improve feature selection for steering.
We evaluate \textsc{NIFS} across multiple SAE-based steering methods and tasks, and demonstrate consistent performance gains over conventional top-$k$ selection.

\end{abstract}
\section{Introduction}

Sparse Autoencoders (SAEs), as an unsupervised method, decompose hidden representations in large language models (LLMs) into more interpretable dimensions, referred to as SAE features \cite{sae,sae1,llamascope,gemmascope}. Recently, researchers have explored steering models in a more controllable manner by manipulating SAE activations and leveraging features that are influential for downstream tasks \cite{axbench,spare,corrsteer}. However, the effectiveness of these approaches critically depends on feature selection, which has not yet been rigorously validated.

Existing steering methods generally follow the ``Contrastive Activation Addition (CAA)'' paradigm \cite{caa}. Given a specific task, model activations are extracted from positive and negative outputs, and a steering vector is computed from the difference between these contrastive activations, which is then added back to the original activations. In SAE-based methods, steering is performed in the SAE feature space. Due to the high dimensionality and sparsity of SAE representations, not all features are relevant to the steering objective, making feature selection a central challenge. Prior work \cite{spare,saif,sta} has explored various statistical feature selection strategies, such as selecting features with the largest differences in mean activations or activation frequencies across contrastive outputs. These approaches can be summarized as statistical top-$k$ selection: assigning each feature a statistical score as a proxy for steering effectiveness and selecting the top-ranked features.

In this work, we investigate existing SAE-based steering methods across multiple generative tasks and show that the widely used statistical top-$k$ selection strategy is suboptimal. First, we examine how steering performance changes with the number of selected features $k$. We find that performance does not monotonically improve with increasing $k$; instead, it often saturates or even degrades. This suggests that additional features can introduce negative steering effects that outweigh the contributions of highly ranked features. Furthermore, by repeatedly sampling feature subsets from top-ranked candidates, we observe that some randomly selected subsets outperform the original top-$k$ selection, providing direct evidence that better feature combinations exist.

We further analyze selected features from a representational perspective by measuring the cosine similarity between SAE features and probing directions derived from linear probes. We observe that similarity decreases rapidly with feature rank: top-ranked features exhibit strong representational alignment, while most lower-ranked features show substantially weaker similarity. However, several low-ranked features still exhibit high similarity, suggesting that statistical ranking alone does not fully capture steering-relevant representations.

One important factor arises from the structure of SAEs themselves. Due to sparsity constraints, SAEs often split a higher-level semantic concept into multiple semantically similar yet functionally differentiated sub-concepts, a phenomenon known as feature splitting \cite{splitting,splitting2}. As a result, some split features may receive high statistical scores, while others remain low-ranked despite possessing substantial steering effectiveness.

Motivated by these observations, we propose NIFS, a plug-and-play enhancement for SAE-based steering methods via \textbf{N}eighbor \textbf{I}ntegrated \textbf{F}eature \textbf{S}election. Standard statistical top-$k$ selection treats SAE features as independent units and can therefore overlook low-ranked but representation-similar features. NIFS first identifies a set of top-ranked core features using existing statistical criteria, and then retrieves neighboring features that are highly similar in representation space. These neighboring features are integrated into the steering vector with similarity-based weights. By aggregating semantically related split features, NIFS recovers fragmented steering signals and achieves more robust and effective steering while maintaining generation quality.

Our findings and contributions are threefold:
\begin{itemize}
  \item We demonstrate that existing statistical top-$k$ selection strategies in SAE-based steering are suboptimal.
  \item We reveal a correlation between the representational similarity of SAE features and their steering effectiveness.
  \item We propose NIFS, a plug-and-play enhancement that integrates representation-similar features to improve steering performance across multiple models and tasks.
\end{itemize}

\section{Preliminaries}
\subsection{Sparse Autoencoders(SAEs).}
SAEs are proposed to disentangle and interpret model hidden representations with a set of concepts by decomposing it into a high dimensional space and then reconstruct them. SAEs consists of two components, encoder and decoder. Encoder projects hidden representation $h \in \mathbb{R}^d$ into SAE space:
\begin{equation}
    Z = \sigma(h\cdot W_{enc} + b_{enc}),
\end{equation}
where SAE activation $Z \in \mathbb{R}^m$ with $m \gg d$, $W_{enc} \in \mathbb{R}^{d \times m}$ and $b_{enc}\in \mathbb{R}^{m}$ represent encoder weight matrix and bias. $\sigma$ is the activation function. Decoder can then reconstruct model representation:
\begin{equation}
    \hat{h}= (Z\cdot W_{dec} + b_{dec}),
\end{equation}
where $\hat{h} \in \mathbb{R} ^d$ is the reconstructed representations, $W_{dec} $, $ b_{dec}$ represent decoder matrix and bias. The SAE encoder and decoder are optimized by reconstruction and sparsity loss:
\begin{equation}
    \mathcal{L}_{SAE}= \lVert h - \hat{h} \rVert_2^2 + \beta \lVert Z \rVert_1.
\end{equation}
Each raw of the decoder matrix $W_{dec}$, denotes as an SAE feature $f \in \mathbb{R}^d$, can be interpreted as a concept.

\begin{figure*}[t]
    \centering

    \begin{subfigure}{0.32\textwidth}
        \centering
        \includegraphics[width=\linewidth]{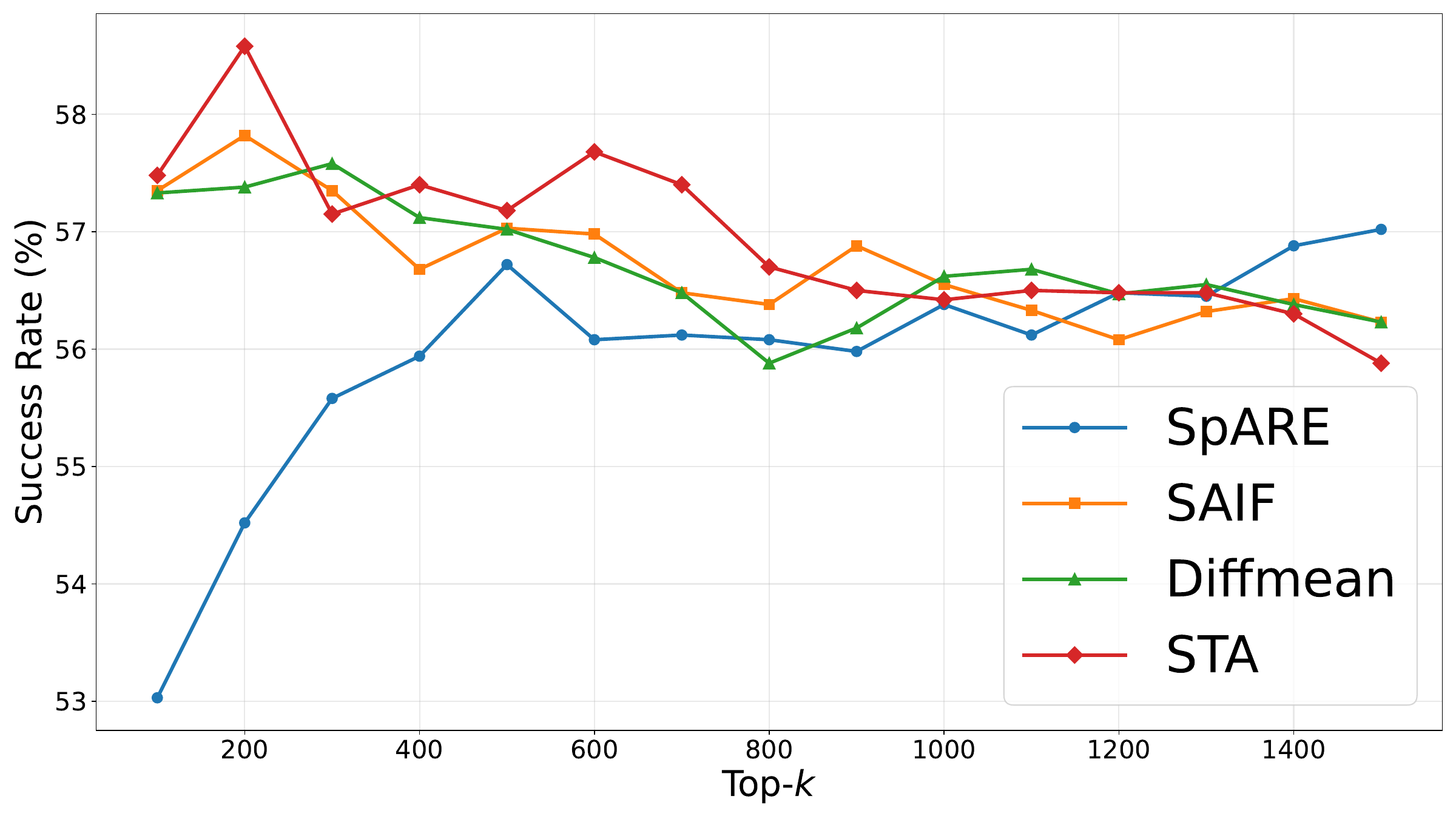}
        \caption{Knowledge Conflicts}
    \end{subfigure}
    \hfill
    \begin{subfigure}{0.32\textwidth}
        \centering
        \includegraphics[width=\linewidth]{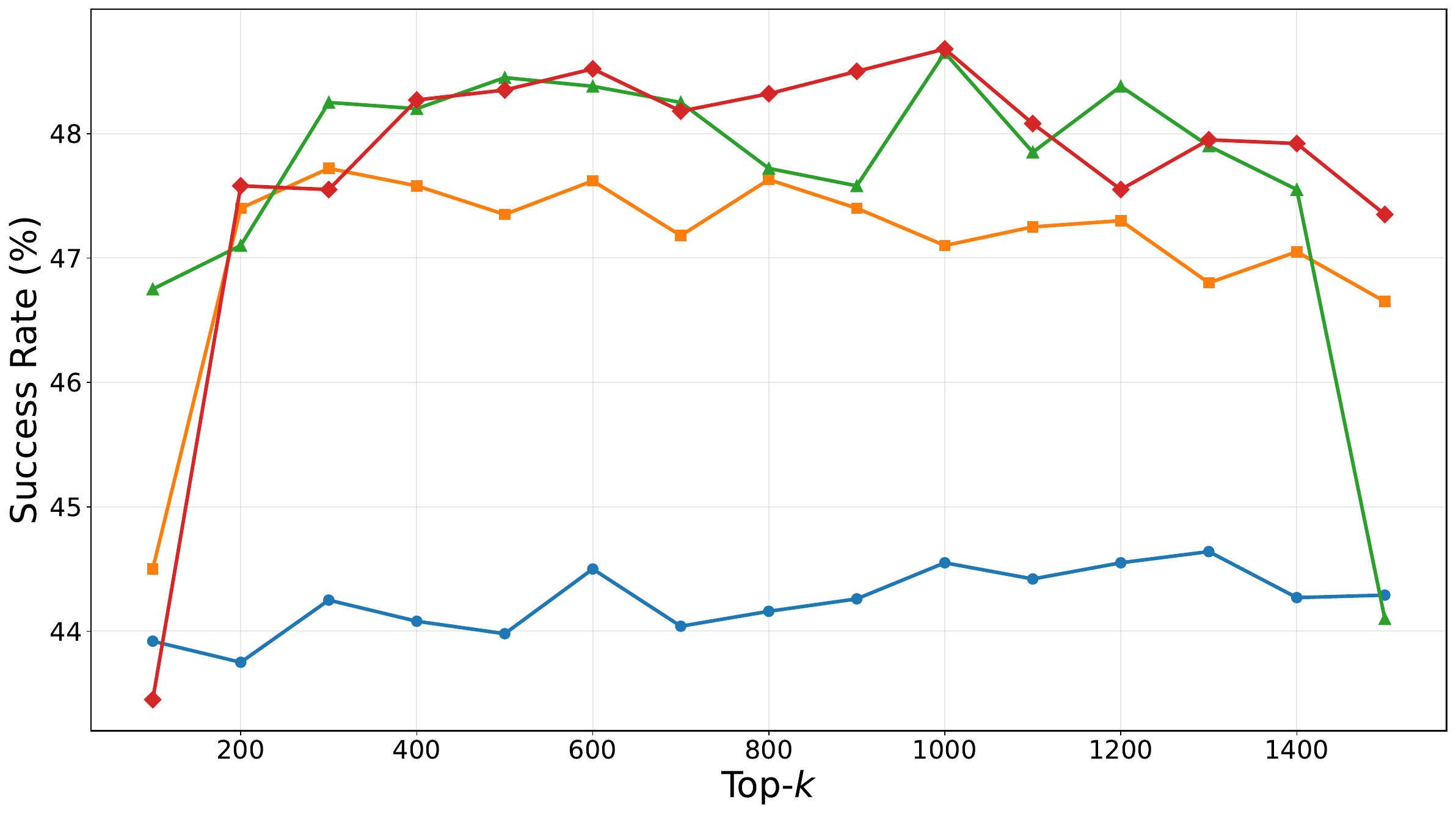}
        \caption{Sentiment}
    \end{subfigure}
    \hfill
    \begin{subfigure}{0.32\textwidth}
        \centering
        \includegraphics[width=\linewidth]{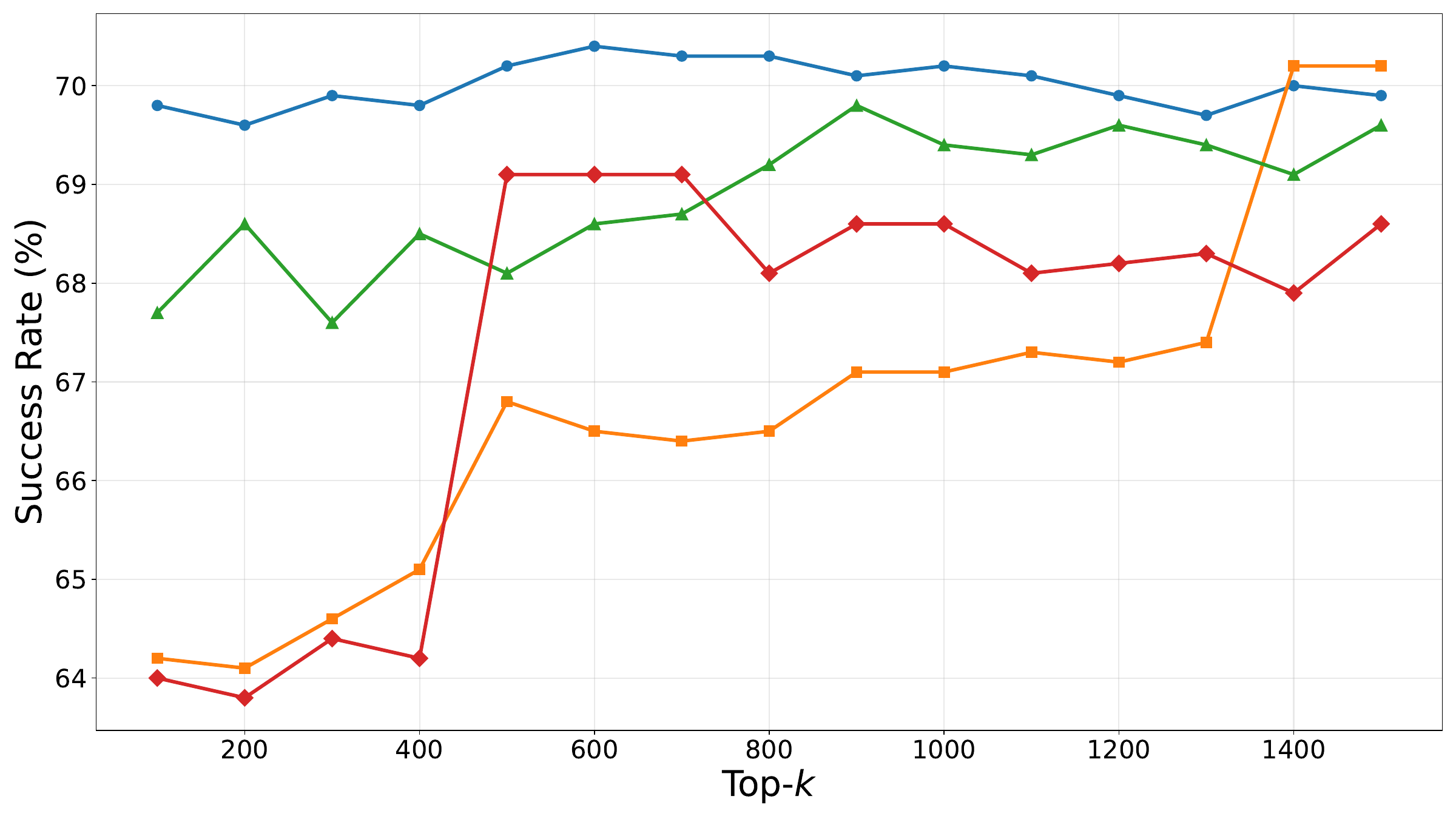}
        \caption{Detoxification}
    \end{subfigure}

    \vspace{6pt}


    \caption{
    Steering success rate on Gemma2-2B across different top-$k$ values ranging from 100 to 1500. The performance is reported for three tasks: (a) Knowledge Conflicts, (b) Sentiment, and (c) Detoxification.
    }
    \label{fig:topk}
\end{figure*}

\begin{figure}
    \centering
    \includegraphics[width=1\linewidth]{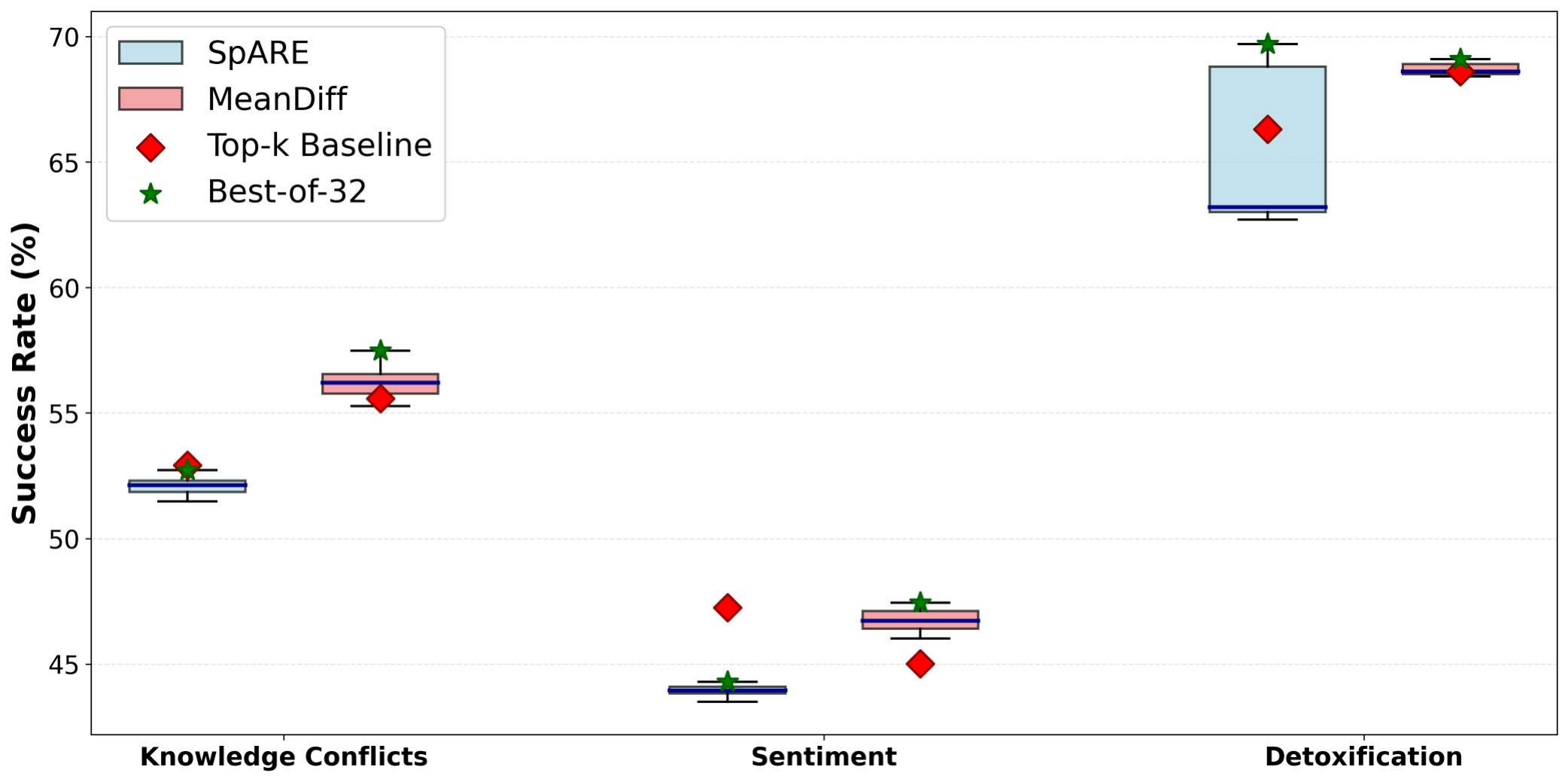}
    \caption{Distribution of success rate over 32 runs of random feature selection on Gemma2-2B.}
    \label{fig:bestof32gemma}
\end{figure}
\subsection{SAE-based Steering}
Steering is a inference time strategy to control model behaviors by modifying model representations. Typically,given an input sequence $x$ and its hidden activation $h \in \mathbb{R}^d$ at a chosen layer, steering modifies $h$ by adding a steering vector $v$:
\begin{equation}
    h' =h + \lambda v,
\end{equation}
where $\lambda$ denotes the steering strength.
Steering strategy, like CAA \cite{caa}, generate $v$ by take the difference in activations between contrastive datasets. Given a dataset $D$ of positive samples $x_{pos}$ and negative samples $x_{neg}$, $v$ can be computed as:
\begin{equation}
    \frac{1}{|D|} \sum _{x_{pos}, x_{neg} \in D}(h(x_{pos})-h(x_{neg})),
\end{equation}
where h($\cdot$) denotes the corresponding model activation of inputs.

Recent works transfer this steering strategy from model space to SAE space, generating $v$ with specific SAE features. Similar to CAA, SAE-based methods first get SAE activations of contrastive inputs $Z_{pos}$, $Z_{neg} \in\mathbb{R} ^m $. A set of prominent features $F$ are then filtered out by designed statistical metrics $S_i(Z_{pos,i},Z_{neg,i})$, which evaluate the importance to steering results of each feature $i \in \{1,...,m\}$ by its activations between contrastive inputs. The final vector $v$ can be computed by features in $F$ and their mean activations:
\begin{equation}
    v = \sum_{f_{pos,j} \in F} \bar{Z}_{pos,j} \cdot f_{pos,j}-\sum_{f_{neg,j} \in F} \bar{Z}_{neg,j} \cdot f_{neg,j}
\end{equation}
where positive and negative features $f_{pos,j}$,$f_{neg,j}$ are classified by their mean activation difference between contrastive samples.

\section{Experimental Setup}
In this section, we introduce the whole experimental setup in this work, including used models, tasks and datasets and traditional statistical top-$k$ selection strategy baselines.

\paragraph{Models.} We conduct our experiments on two base LLMs: Gemma-2-2B and Llama-2-7B and their corresponding SAEs.
Details can be found in Appendix \ref{subsec:llmsae}.

\paragraph{Tasks, Datasets, and Evaluation Metrics.}
We evaluate our method on three open-ended steering tasks: Knowledge Conflicts, Sentiment, and Detoxification. Detailed descriptions of datasets, evaluation settings, and metrics are provided in Appendix~\ref{app:task_eval_details}.

\paragraph{Baselines.}
\label{baseline}
We select four representative statistical selection strategies: 1) Diffmean, selecting features that differ most in mean activations. 2) SAIF \cite{saif}, selecting features that differ most in activated frequency. 3) STA \cite{sta}, filtering out features that both differ in amplitude and frequency. It filters out as the intersection of results of mean-diff and freq-diff. 4) SpARE \cite{spare}, selecting features based on the mutual information between their activations and model behaviors.


\section{Understanding Statistical Selection Strategies in SAE-based Steering}
\subsection{Overview}

Feature selection is a critical component of SAE-based steering. The objective of feature selection is to identify features that are most effective in steering the model toward target behaviors. To this end, prior work proposes various statistical metrics to quantify the correlation between model behaviors and feature activations, and selects the top-$k$ features with the highest scores. However, a fundamental question remains: \textit{do these statistical top-$k$ selection strategies indeed identify the most effective features for steering?}

To investigate this question, we conduct empirical analyses across multiple steering methods and tasks. The results consistently indicate that selecting features solely based on statistical top-$k$ scores is suboptimal for steering. Furthermore, inspired by the feature splitting phenomenon \cite{splitting, splitting2}, we analyze the representational similarity between top-$k$ features and probing direction. Our findings show that features with high representational similarity to the probing direction tend to exhibit comparable steering effectiveness, suggesting that incorporating split features can improve existing statistical feature selection strategies.
\begin{figure*}
    \centering
    \includegraphics[width=1\linewidth]{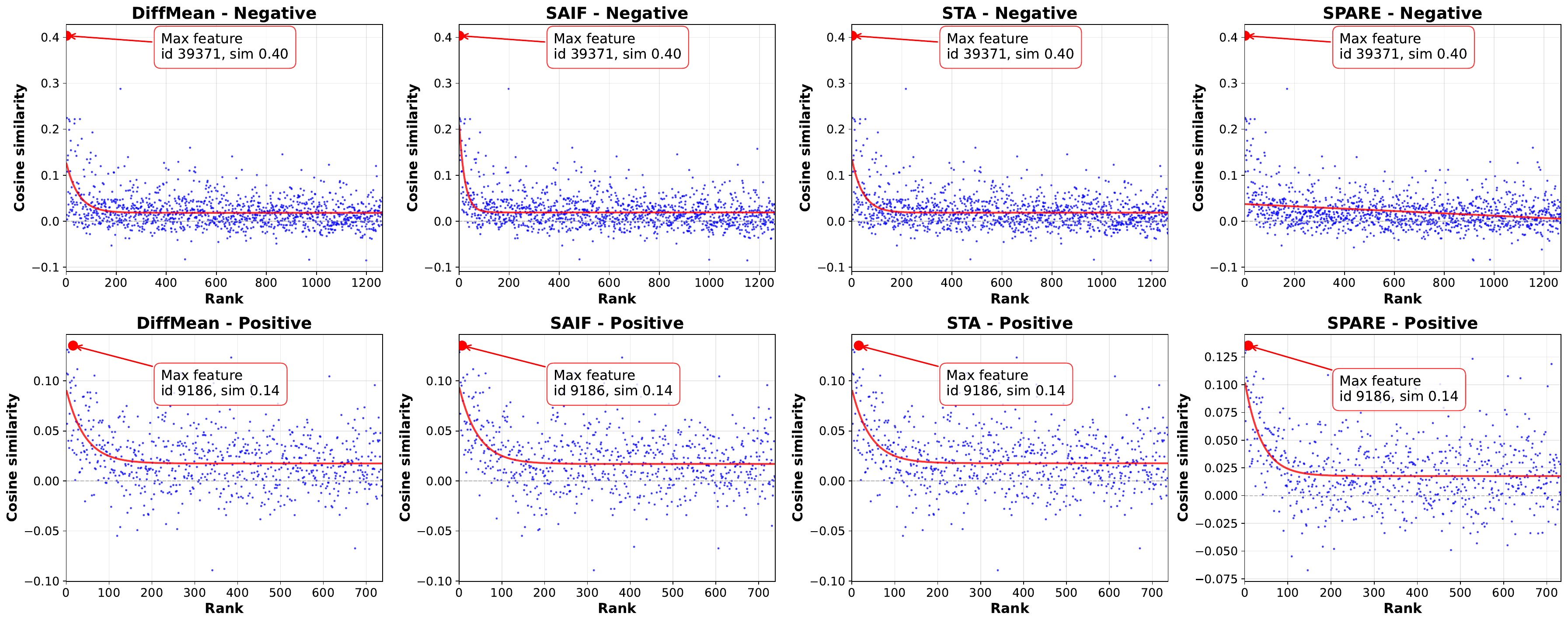}
    \caption{Cosine similarity between SAE features and probing directions across feature ranks for the Sentiment task on Gemma2-2B.
Features are ranked by their importance scores under different methods.
Blue points correspond to individual features, while the red line indicates an exponential fit trend.
Negative and positive probing directions are shown in separate rows.
}
    \label{fig:probe-gemma2-senti}
\end{figure*}
\subsection{Top-k selected features are suboptimal}
\label{topksub}
We begin by evaluating the performance of different SAE-based steering methods across a range of $k$ values. Specifically, we measure the success rate (SR) of baseline methods on three tasks with $k \in [100, 1500]$ and a step size of 100. The results of Gemma2-2B are shown in Figure~\ref{fig:topk}. The results of Llama2-7B are shown in Figure~\ref{fig:topk-llama}.
In general, as $k$ increases, the performance of most methods rises rapidly at first and then enters a relatively stable regime.
Across all methods, we find that none exhibits steering performance that consistently improves with increasing $k$. This observation suggests that not all top-$k$ selected features, or naive combinations of top-$k$ features, possess steering effectiveness proportional to their ranking scores. As $k$ increases beyond a certain point, features with little or even negative steering effect begin to outweigh those with positive contributions, leading to performance saturation or degradation. This phenomenon provides indirect evidence that statistical top-$k$ feature selection does not always identify the most effective features for steering.

To further validate this observation, we compare top-$k$ feature selection with a randomized selection baseline. Specifically, we construct a hybrid feature set by randomly sampling $0.5k$ features from those ranked between $0.5k$ and $1.5k$, and combining them with the top-$0.5k$ features. In this experiment, we fix $k = 100$.
We repeat the experiment 32 times and report the performance distribution of Gemma2-2B in Figure~\ref{fig:bestof32gemma}. The results show that some randomized subsets outperform that of the original top-$k$ selection, indicating the existence of feature subsets with higher steering effectiveness than those identified by statistical ranking.

\subsection{Steering Effectiveness Correlates with Representation Similarity}
\label{probesec}

To further understand the steering effectiveness of SAE features, we train linear probes using the same activations employed for SAE feature selection and analyze the representational relationship between selected features and the corresponding probing directions. Specifically, we compute the cosine similarity between each feature and its corresponding positive or negative probing direction.

Figure~\ref{fig:probe-gemma2-senti} visualizes the results for the Sentiment task on Gemma2-2B, while additional results are provided in the Appendix. Across methods, cosine similarity exhibits an approximately exponential decay with feature rank, with highly similar features concentrated among top-ranked features. Moreover, the features with the highest cosine similarity are consistently selected among the top-ranked features by different methods, indicating a strong correlation between representational similarity and steering effectiveness.

At the same time, several low-ranked features still exhibit high similarity to the probing directions. This phenomenon can be explained by feature splitting, where SAEs decompose a higher-level semantic concept into multiple semantically similar yet distinguishable sub-concepts. Some split features receive high statistical ranks due to stronger activation responses, while others remain low-ranked despite possessing substantial steering effectiveness. These observations suggest that integrating representation-similar split features may improve existing statistical feature selection strategies.

\begin{algorithm}[tb]
\caption{Neighbor Integration for NIFS}
\label{alg:NIFS}
\DontPrintSemicolon

\KwIn{Statistical rank $R$; feature set size $k$; pool size $\alpha$; neighbor size $\beta$; steering activation $\bar{Z}$}
\KwOut{Enhanced feature set $\boldsymbol{F}'_k$; enhanced steering activation $\bar{Z}'$}

\textbf{Stage 1: Neighbor Integration}\;

Generate pool set $\boldsymbol{P}_\alpha = \{ f_i \mid R_i \le \alpha k \}$\;

\For{each feature $f_i \in \boldsymbol{P}_\alpha$}{
    Add $f_i$ to core set $\boldsymbol{C}_\beta$\;

    Compute top-$\beta$ similar neighbors of $f_i$\;

    $\boldsymbol{N}^i \leftarrow \mathrm{KNN}(f_i, \beta, \boldsymbol{P}_\alpha)$\;

    $\boldsymbol{N}_\beta \leftarrow \boldsymbol{N}_\beta \cup \boldsymbol{N}^i$\;

    \If{$|\boldsymbol{C}_\beta| + |\boldsymbol{N}_\beta| = k$}{
        break\;
    }
}

\textbf{Stage 2: Feature Reweighting}\;

\For{each feature $f_i \in \boldsymbol{N}_\beta$}{
    $\bar{Z}'_i \leftarrow \bar{Z}_i \cdot 
    \max\limits_{f_j \in \boldsymbol{C}_\beta} \mathrm{Sim}(f_i, f_j)$\;
}

$\boldsymbol{F}'_k \leftarrow \boldsymbol{C}_\beta \cup \boldsymbol{N}_\beta$\;

\end{algorithm}

\section{NIFS: Neighbor Integrated Feature Selection}
In this section, we introduce Neighbor Integrated Feature Selection (NIFS), a plug-and-play feature selection strategy designed to improve SAE-based steering methods that rely on statistical feature selection.
Motivated by our empirical findings that effective steering features often reside in representationally adjacent groups due to feature splitting, \textsc{NIFS} augments conventional score-based selection by systematically incorporating semantically similar neighboring features.

Rather than discarding statistical scores, \textsc{NIFS} treats them as a coarse but informative signal and refines the selection through representation-aware expansion and controlled reweighting.
The method consists of two stages: \textit{Neighbor Integration} and \textit{Feature Reweighting}.
The full pipeline is summarized in Algorithm~\ref{alg:NIFS}.

\subsection{Neighbor Integration}
We consider a generic SAE-based steering method that assigns each feature $f_i$ a statistical score $S_i$ and a corresponding rank $R_i$.
A standard top-$k$ strategy selects the feature set
\[
F_k = \{ f_i \mid R_i \le k \}.
\]

As a working assumption, we observe that statistical scores are informative in a relative sense: features with higher scores are, on average, more likely to be effective for steering.
However, due to feature splitting, effective features may be distributed across semantically similar but individually low-scoring components.
To mitigate this issue, we first expand the candidate space by constructing a \textit{pool set}
\[
P_{\alpha} = F_{\alpha k},
\]
where $\alpha > 1$ controls the pool size.
This step aims to retain most effective features while still anchoring the selection to score-based ranking.

Next, we decompose the final selection into two parts: a \textit{core set} $\boldsymbol{C}_{\beta}$ and a \textit{neighbor set} $\boldsymbol{N}_{\beta}$.
We traverse features in $P_{\alpha}$ in descending order of rank.
High-ranked features are preferentially added to the core set, reflecting their strong statistical evidence.
For each core feature, we then retrieve its top-$\beta$ nearest neighbors within $P_{\alpha}$ according to representation similarity, measured by cosine similarity between SAE feature vectors.
Nearest neighbors are identified using a $k$-nearest neighbor (KNN) search and added to the neighbor set.

This process continues until the total number of selected features satisfies
\[
|\boldsymbol{C}_{\beta}| + |\boldsymbol{N}_{\beta}| = k.
\]
Through this mechanism, \textsc{NIFS} explicitly recovers low-scoring but representationally adjacent features that are likely to share steering functionality with core features.

\begin{table*}[t]
\centering
\scriptsize
\renewcommand{\arraystretch}{0.95}
\setlength{\aboverulesep}{1pt}
\setlength{\belowrulesep}{1pt}
\setlength{\tabcolsep}{6pt}
\caption{Comparison of original and NIFS-enhanced SAE-based baselines across models and tasks. SR denotes the success rate and Flu. denotes the n-gram fluency.}
\label{tab:main-exp}
\resizebox{\textwidth}{!}{%
\begin{tabular}{l l c c c c c c}
\toprule
\multirow{2}{*}{Model}
& \multirow{2}{*}{Method}
& \multicolumn{1}{c}{\textbf{KC}}
& \multicolumn{2}{c}{\textbf{Sentiment}}
& \multicolumn{2}{c}{\textbf{Detoxification}} \\
\cmidrule(lr){3-3}
\cmidrule(lr){4-5}
\cmidrule(lr){6-7}
& & SR & SR & Flu. & SR & Flu. \\
\midrule

\multirow{11}{*}{\textbf{Llama2-7B}}
& Vanilla
& $38.14 {\scriptstyle\pm 3.97}$
& $38.16 {\scriptstyle\pm 1.31}$ & $4.43 {\scriptstyle\pm 0.18}$
& $62.79 {\scriptstyle\pm 0.81}$ & $3.92 {\scriptstyle\pm 0.28}$ \\

& CAA
& $63.61 {\scriptstyle\pm 3.56}$
& $44.95 {\scriptstyle\pm 0.99}$ & $4.37 {\scriptstyle\pm 0.13}$
& $67.94 {\scriptstyle\pm 1.00}$ & $3.92 {\scriptstyle\pm 0.27}$ \\

& Probe
& $65.38 {\scriptstyle\pm 1.55}$
& $47.56 {\scriptstyle\pm 1.18}$ & $4.30 {\scriptstyle\pm 0.04}$
& $71.40 {\scriptstyle\pm 0.90}$ & $3.80 {\scriptstyle\pm 0.01}$ \\

\cmidrule(lr){2-7}

& Diffmean
& $68.49 {\scriptstyle\pm 3.04}$
& $59.67 {\scriptstyle\pm 0.95}$ & $3.98 {\scriptstyle\pm 0.17}$
& $80.80 {\scriptstyle\pm 1.06}$ & $3.83 {\scriptstyle\pm 0.01}$ \\

& \textbf{+NIFS}
& $\textbf{68.57} {\scriptstyle\pm 0.88}$
& $\textbf{62.69*} {\scriptstyle\pm 1.29}$ & $3.95 {\scriptstyle\pm 0.04}$
& $\textbf{85.30*} {\scriptstyle\pm 0.86}$ & $3.84 {\scriptstyle\pm 0.01}$ \\

& SAIF
& $68.93 {\scriptstyle\pm 3.32}$
& $47.42 {\scriptstyle\pm 1.24}$ & $4.17 {\scriptstyle\pm 0.18}$
& $73.10 {\scriptstyle\pm 0.49}$ & $3.82 {\scriptstyle\pm 0.01}$ \\

& \textbf{+NIFS}
& $\textbf{70.72*} {\scriptstyle\pm 1.69}$
& $\textbf{48.72*} {\scriptstyle\pm 0.71}$ & $4.15 {\scriptstyle\pm 0.09}$
& $\textbf{74.63*} {\scriptstyle\pm 1.00}$ & $3.82 {\scriptstyle\pm 0.01}$ \\

& STA
& $69.97 {\scriptstyle\pm 3.50}$
& $45.11 {\scriptstyle\pm 0.90}$ & $4.21 {\scriptstyle\pm 0.13}$
& $72.27 {\scriptstyle\pm 0.71}$ & $3.82 {\scriptstyle\pm 0.01}$ \\

& \textbf{+NIFS}
& $\textbf{71.16} {\scriptstyle\pm 1.71}$
& $\textbf{47.26*} {\scriptstyle\pm 1.12}$ & $4.19 {\scriptstyle\pm 0.09}$
& $\textbf{74.48*} {\scriptstyle\pm 1.09}$ & $3.82 {\scriptstyle\pm 0.01}$ \\

& SpARE
& $69.71 {\scriptstyle\pm 3.19}$
& $65.63 {\scriptstyle\pm 1.41}$ & $4.26 {\scriptstyle\pm 0.13}$
& $76.55 {\scriptstyle\pm 0.62}$ & $3.81 {\scriptstyle\pm 0.00}$ \\

& \textbf{+NIFS}
& $\textbf{71.34*} {\scriptstyle\pm 1.78}$
& $\textbf{66.94} {\scriptstyle\pm 1.17}$ & $4.26 {\scriptstyle\pm 0.14}$
& $\textbf{77.83*} {\scriptstyle\pm 0.68}$ & $3.83 {\scriptstyle\pm 0.01}$ \\

\midrule

\multirow{11}{*}{\textbf{Gemma2-2B}}
& Vanilla
& $41.07 {\scriptstyle\pm 2.71}$
& $38.53 {\scriptstyle\pm 0.89}$ & $4.65 {\scriptstyle\pm 0.22}$
& $62.54 {\scriptstyle\pm 1.90}$ & $4.11 {\scriptstyle\pm 0.26}$ \\

& CAA
& $60.34 {\scriptstyle\pm 1.43}$
& $42.48 {\scriptstyle\pm 1.25}$ & $4.52 {\scriptstyle\pm 0.13}$
& $70.78 {\scriptstyle\pm 0.57}$ & $4.11 {\scriptstyle\pm 0.26}$ \\

& Probe
& $42.41 {\scriptstyle\pm 2.89}$
& $44.79 {\scriptstyle\pm 1.61}$ & $4.53 {\scriptstyle\pm 0.11}$
& $63.49 {\scriptstyle\pm 1.39}$ & $4.00 {\scriptstyle\pm 0.01}$ \\

\cmidrule(lr){2-7}

& Diffmean
& $58.47 {\scriptstyle\pm 1.92}$
& $46.86 {\scriptstyle\pm 1.03}$ & $4.44 {\scriptstyle\pm 0.06}$
& $73.77 {\scriptstyle\pm 0.90}$ & $4.00 {\scriptstyle\pm 0.01}$ \\

& \textbf{+NIFS}
& $\textbf{59.66*} {\scriptstyle\pm 1.78}$
& $\textbf{47.47*} {\scriptstyle\pm 0.90}$ & $4.43 {\scriptstyle\pm 0.06}$
& $\textbf{73.79} {\scriptstyle\pm 0.84}$ & $4.00 {\scriptstyle\pm 0.01}$ \\

& SAIF
& $58.87 {\scriptstyle\pm 1.41}$
& $47.60 {\scriptstyle\pm 1.01}$ & $4.46 {\scriptstyle\pm 0.10}$
& $72.86 {\scriptstyle\pm 0.49}$ & $4.00 {\scriptstyle\pm 0.01}$ \\

& \textbf{+NIFS}
& $\textbf{59.54} {\scriptstyle\pm 1.94}$
& $\textbf{48.39*} {\scriptstyle\pm 0.53}$ & $4.48 {\scriptstyle\pm 0.08}$
& $\textbf{73.32*} {\scriptstyle\pm 0.99}$ & $4.00 {\scriptstyle\pm 0.01}$ \\

& STA
& $58.91 {\scriptstyle\pm 1.59}$
& $49.97 {\scriptstyle\pm 1.44}$ & $4.51 {\scriptstyle\pm 0.21}$
& $73.66 {\scriptstyle\pm 1.53}$ & $4.00 {\scriptstyle\pm 0.01}$ \\

& \textbf{+NIFS}
& $\textbf{60.04*} {\scriptstyle\pm 2.04}$
& $\textbf{50.54} {\scriptstyle\pm 1.33}$ & $4.50 {\scriptstyle\pm 0.16}$
& $\textbf{74.36*} {\scriptstyle\pm 1.35}$ & $4.01 {\scriptstyle\pm 0.01}$ \\

& SpARE
& $59.79 {\scriptstyle\pm 1.25}$
& $44.36 {\scriptstyle\pm 1.05}$ & $4.45 {\scriptstyle\pm 0.04}$
& $71.27 {\scriptstyle\pm 0.75}$ & $4.00 {\scriptstyle\pm 0.01}$ \\

& \textbf{+NIFS}
& $\textbf{60.09} {\scriptstyle\pm 1.33}$
& $\textbf{46.35*} {\scriptstyle\pm 0.92}$ & $4.49 {\scriptstyle\pm 0.11}$
& $\textbf{75.48*} {\scriptstyle\pm 0.76}$ & $4.01 {\scriptstyle\pm 0.01}$ \\

\bottomrule
\end{tabular}
}
\end{table*}
\subsection{Feature Reweighting}
While neighbor integration helps recover semantically relevant features, directly assigning them the same steering strength as core features may introduce noise, as neighbor features often exhibit lower input sensitivity.
To balance coverage and robustness, we apply a feature reweighting scheme that modulates the steering intensity of neighbor features based on their similarity to the core set.

Specifically, for each neighbor feature $f_i \in \boldsymbol{N}_{\beta}$, we rescale its original steering coefficient $\bar{Z}_i$ as
\[
\bar{Z}'_i = \bar{Z}_i \cdot \max_{f_j \in \boldsymbol{C}_{\beta}} \mathrm{Sim}(f_i, f_j),
\]
where $\mathrm{Sim}(\cdot, \cdot)$ denotes cosine similarity between feature representations.

This reweighting strategy ensures that neighbor features with stronger semantic alignment to core features exert greater influence, while suppressing spurious neighbors.
As a result, \textsc{NIFS} enhances steering effectiveness by integrating representational structure without sacrificing the stability of score-based selection.

\section{Experiments}
In this section, we conduct a series of experiments to evaluate the effectiveness of NIFS by addressing the following research questions (RQs):

\begin{itemize}
    \item RQ1: How does NIFS enhance baselines top-$k$ selection strategies? (Section \ref{subsec:main})
    \item RQ2: How does the neighbor size affect the selected feature similarity relationships and steering performance? (Section \ref{subsec:nei})
    \item RQ3: How do the individual components of NIFS contribute to its overall effectiveness? (Section \ref{subsec:abl})
    \item RQ4: How interpretable is NIFS, and can we understand its feature selection behavior? (Section \ref{subsec:int})
\end{itemize}

\subsection{Setup}
In addition to the four SAE-based baselines introduced in Section~\ref{baseline}, we compare our NIFS-enhanced methods with two non-SAE steering methods: 1) CAA, which uses the mean difference between contrastive dense activations as the steering vector; and 2) Probe, which uses the probing directions trained in Section~\ref{probesec} as steering vectors. The hyperparameter details are in Appendix \ref{subsec:hyper}.

\subsection{Overall Performance Comparison}
\label{subsec:main}

we repeated all experiments using five different seeds and reported the mean and standard deviation in Table~\ref{tab:main-exp}.
In addition, we conduct paired t-tests, where results marked with * indicate statistical significance (p < 0.1).

From the results, we make the following three key observations.
First, NIFS consistently improves steering performance across all evaluated SAE-based baselines, regardless of their original effectiveness. For example, on Llama2-7B, applying NIFS to the STA strategy improves the steering success rate by 2.15\% on Sentiment and by 2.21\% on Detoxification. Similar improvements are observed across other baselines and on Gemma2-2B, indicating that NIFS provides a generally strong enhancement rather than benefiting only weak methods.

Second, the performance gains brought by NIFS do not come at the expense of generation quality, even on more complex generation tasks. Across both sentiment control and detoxification, fluency remains largely stable after applying NIFS, with variations consistently within 0.1. This demonstrates that NIFS improves steering effectiveness while preserving generation quality, which is important for constrained generation settings.

Third, NIFS yields more pronounced gains when applied to already strong baselines, especially on challenging tasks. For instance, on Llama2-7B detoxification, SPARE achieves a relatively high success rate of 80.80, which is further improved to 85.30 (+4.50) after incorporating NIFS. Similar trends are observed for STA and SAIF across multiple tasks. This suggests that while statistical selection strategies can serve as effective coarse filters, NIFS further refines the selected feature set by identifying more informative and synergistic features, leading to additional gains even when the initial selection quality is high.

\begin{figure}[t]
    \centering
    \begin{subfigure}{0.48\linewidth}
        \centering
        \includegraphics[width=\linewidth]{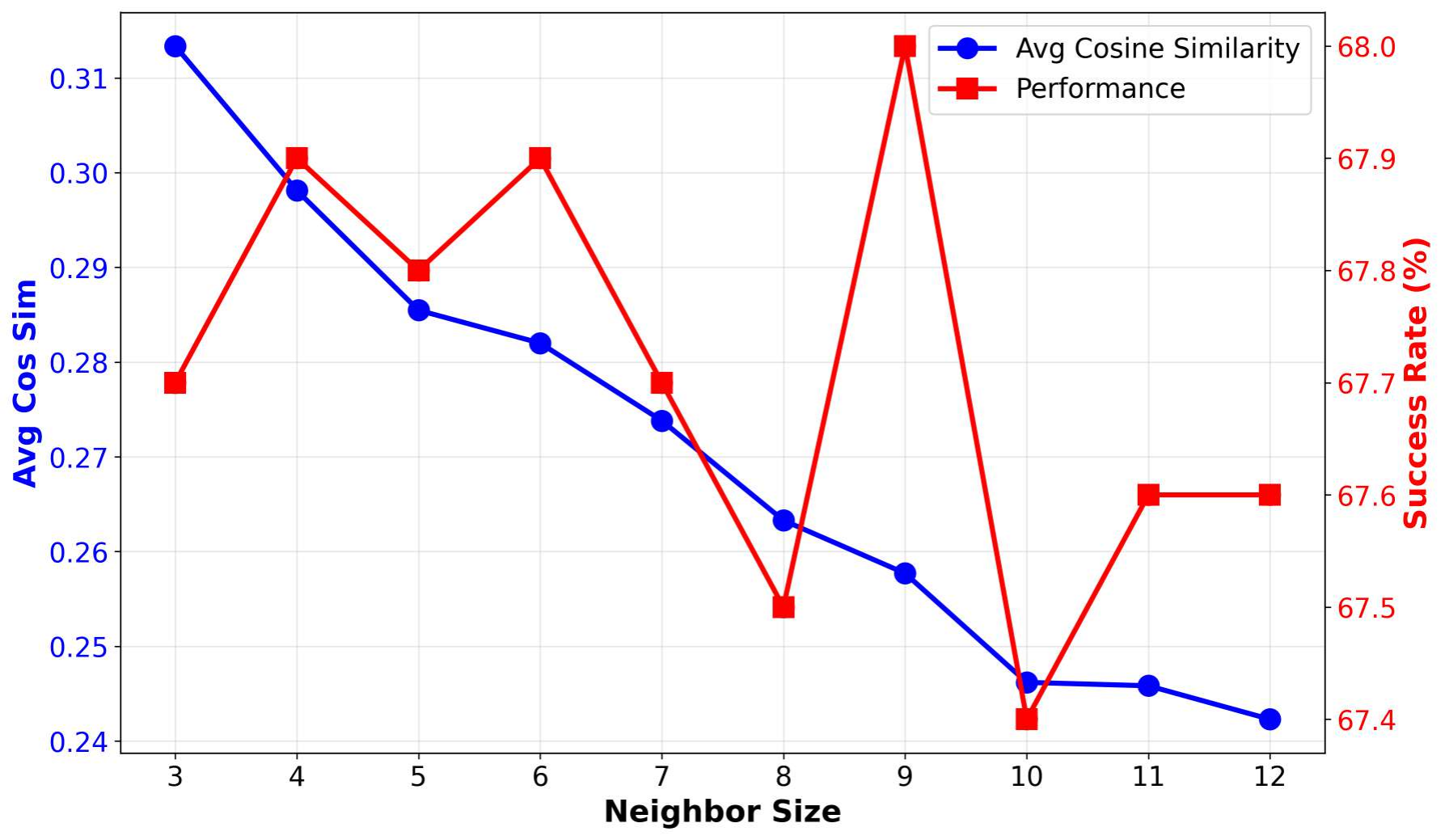}
        \caption{SPARE}
    \end{subfigure}
    \hfill
    \begin{subfigure}{0.48\linewidth}
        \centering
        \includegraphics[width=\linewidth]{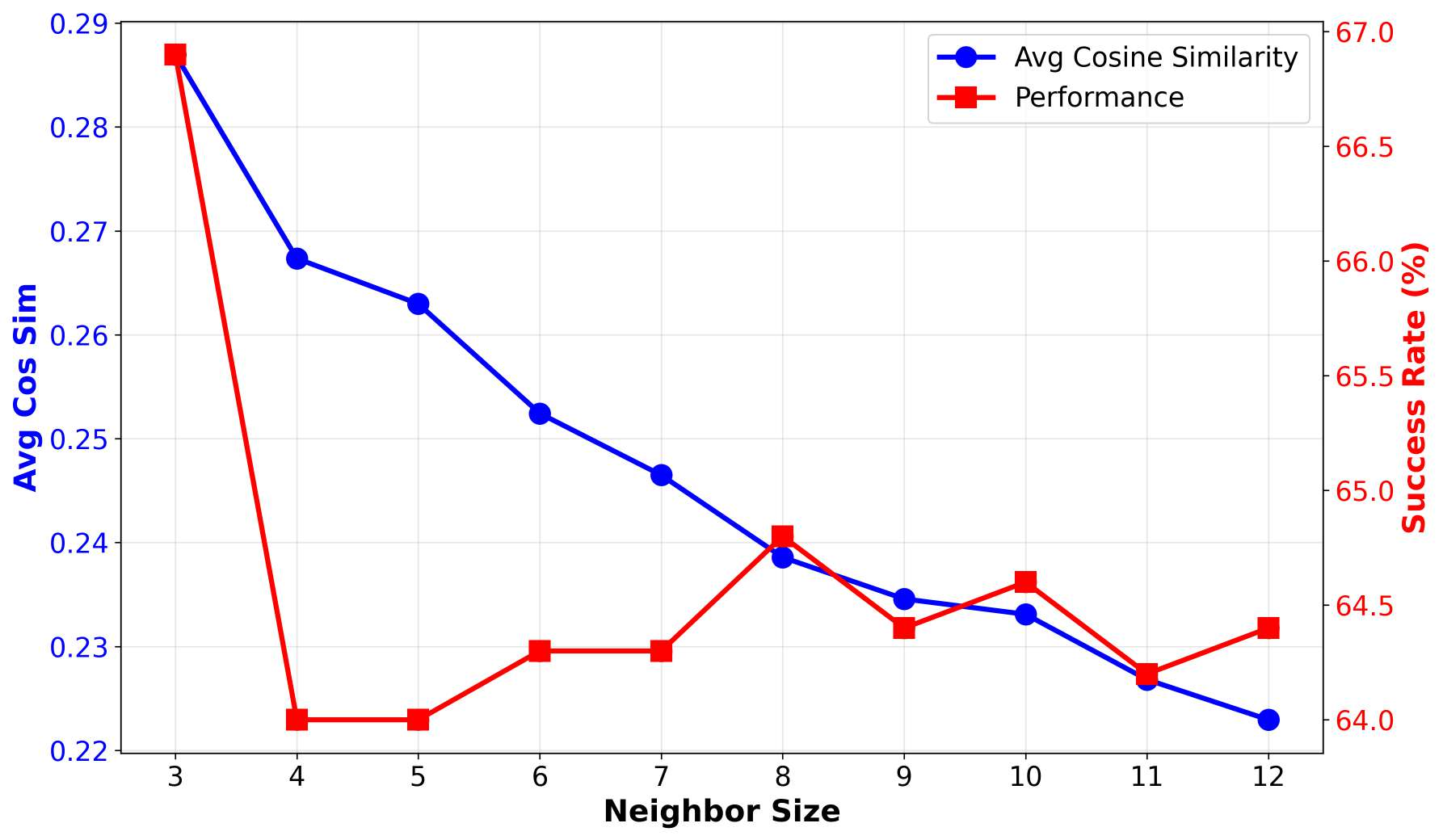}
        \caption{SAIF}
    \end{subfigure}

    \vspace{4pt}

    \begin{subfigure}{0.48\linewidth}
        \centering
        \includegraphics[width=\linewidth]{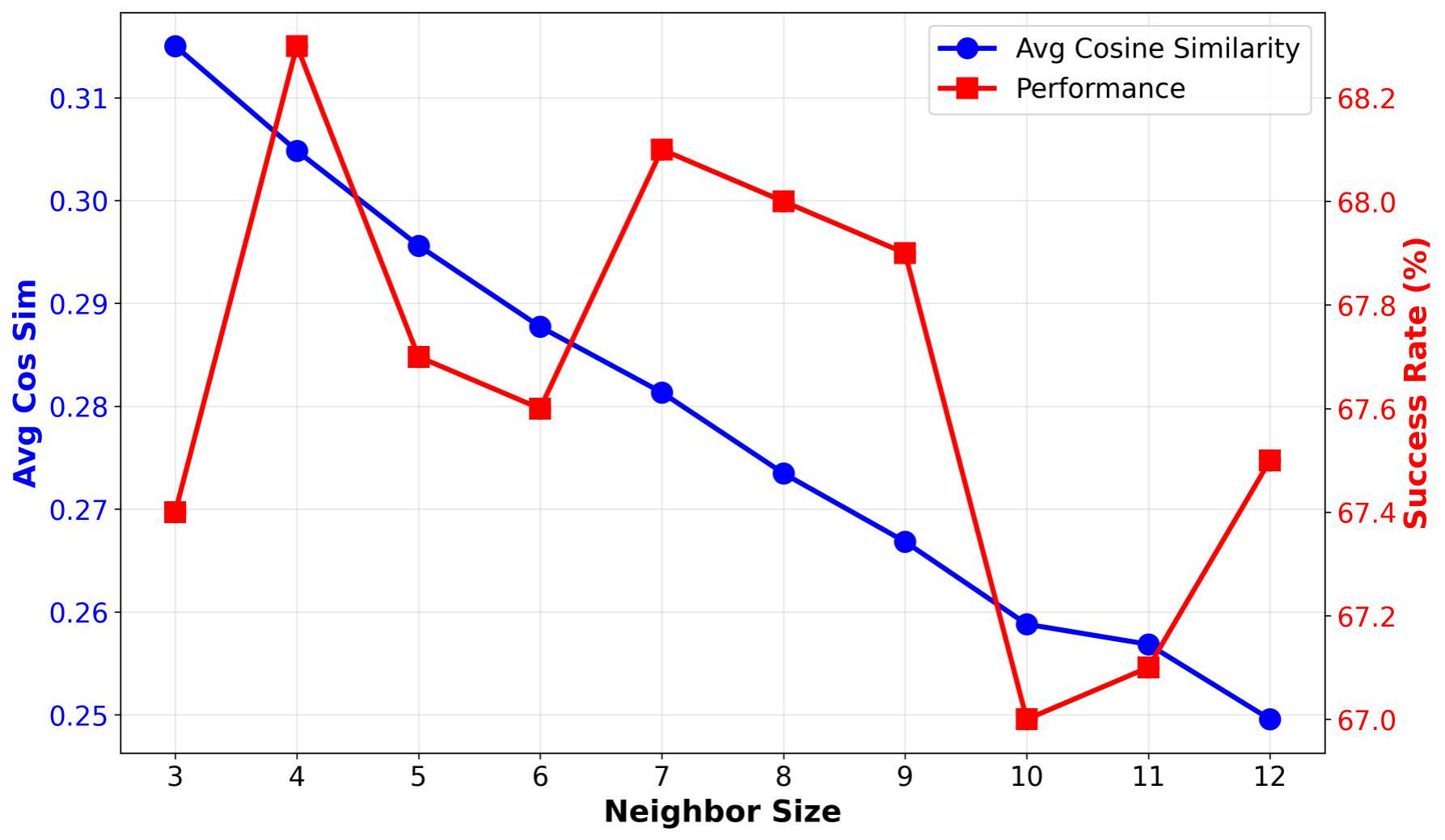}
        \caption{STA}
    \end{subfigure}
    \hfill
    \begin{subfigure}{0.48\linewidth}
        \centering
        \includegraphics[width=\linewidth]{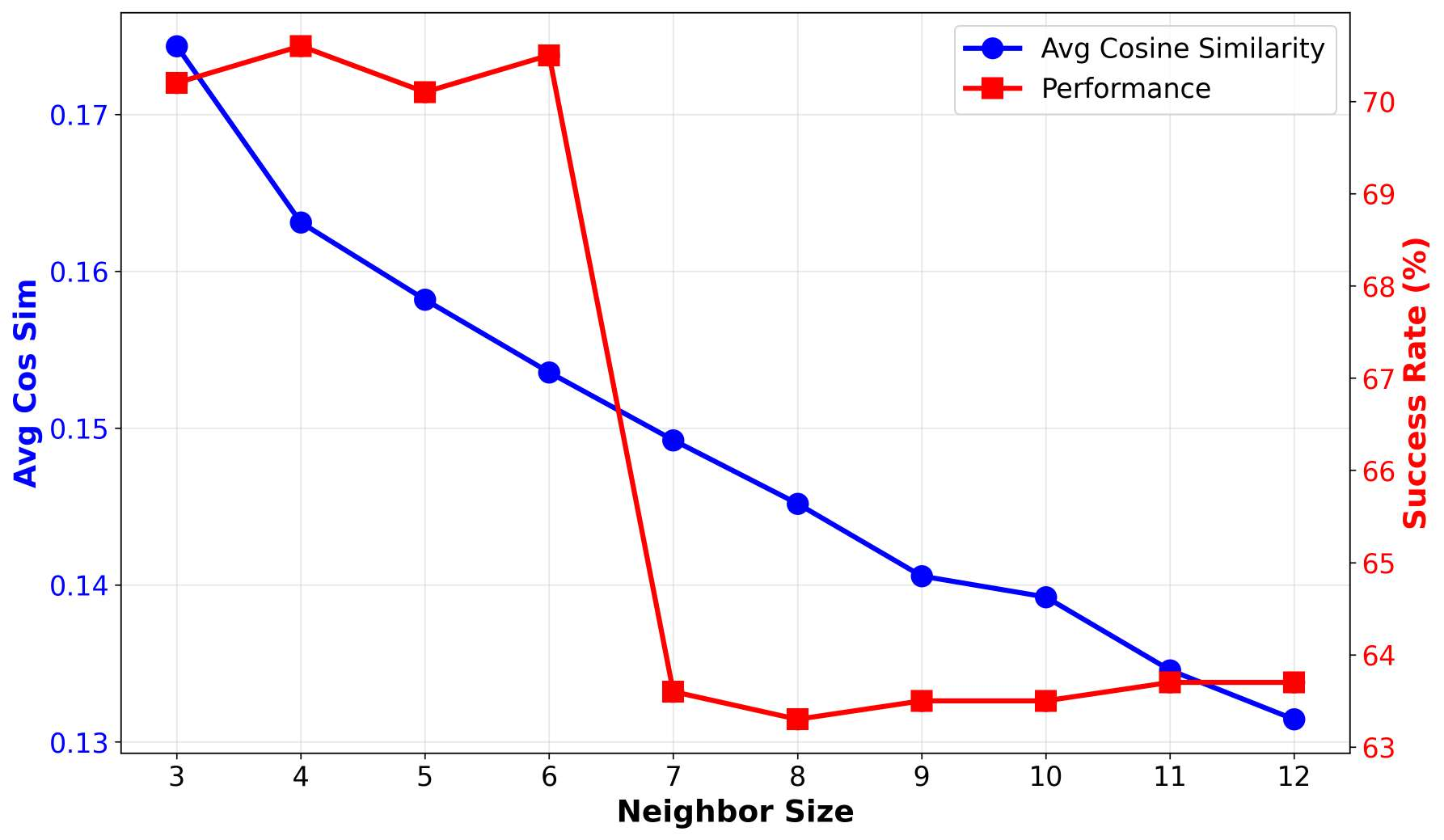}
        \caption{Diffmean}
    \end{subfigure}

    \caption{
    Steering success rate (SR) and average cosine similarity of neighbor features on Gemma2-2B for the Detoxification task across different neighbor sizes (top-$k$). The four subfigures correspond to the four methods: SPARE, SAIF, STA, and Diffmean.
    }
    \label{fig:gemma-detox-neighb}
\end{figure}

\subsection{Generalization to Larger Models}
\begin{table}[t]
\centering
\small
\caption{Results on Gemma3-12B. SR denotes the success rate and Flu. denotes the n-gram fluency.}
\label{tab:gemma3-12b}
\setlength{\tabcolsep}{6pt}
\begin{tabular}{l c c c c c}
\toprule
\textbf{Methods}
& \textbf{KC}
& \multicolumn{2}{c}{\textbf{Detox}}
& \multicolumn{2}{c}{\textbf{Sentiment}} \\
\cmidrule(lr){2-2}
\cmidrule(lr){3-4}
\cmidrule(lr){5-6}
& SR
& SR & Flu.
& SR & Flu. \\
\midrule

Diffmean
& 63.03
& 70.98 & 3.957
& 60.82 & 4.065 \\

\textbf{+NIFS}
& \textbf{63.98}
& \textbf{71.23} & 3.963
& \textbf{63.57} & 4.047 \\

\midrule

SAIF
& 62.53
& 73.64 & 3.948
& 59.21 & 4.258 \\

\textbf{+NIFS}
& \textbf{62.78}
& \textbf{73.89} & 3.938
& \textbf{60.46} & 4.203 \\

\midrule

STA
& 62.68
& 70.89 & 3.948
& 56.73 & 4.271 \\

\textbf{+NIFS}
& \textbf{63.78}
& \textbf{71.81} & 3.947
& \textbf{58.86} & 4.252 \\

\midrule

SpARE
& 61.33
& 75.15 & 3.861
& \textbf{67.18} & 4.186 \\

\textbf{+NIFS}
& \textbf{62.88}
& \textbf{75.31} & 3.890
& 67.01 & 4.178 \\

\bottomrule
\end{tabular}
\end{table}
To further evaluate the scalability of NIFS, we extend our experiments to the larger Gemma3-12B model. We apply NIFS to the same SAE-based steering baselines and evaluate its effectiveness across the steering tasks.

The results are presented in Table~\ref{tab:gemma3-12b}. NIFS improves the steering performance of existing feature selection strategies on Gemma3-12B in most settings, demonstrating that the benefits of integrating representation-similar neighboring features are not limited to smaller models. These results further support the generalizability of NIFS across model scales.

\subsection{Effects of Neighbor Size}
\label{subsec:nei}
In NIFS, the neighbor size $\beta$ controls the degree of feature integration by determining how many neighboring features are incorporated for each core feature. To investigate how the integration strength affects feature selection and the final steering performance, we vary $\beta$ in the range of $[3,12]$ and examine both the average cosine similarity between each core feature and its selected neighbors, as well as the corresponding steering performance. We report the results on the Gemma2-2B detoxification task in Figure~\ref{fig:gemma-detox-neighb}.

Following~\cite{llamascope}, we contextualize the observed feature similarity using a principled baseline derived from the Johnson--Lindenstrauss lemma in its inner-product form. Specifically, if $F$ features are randomly projected into a $D$-dimensional space, the probability that any pair among $\binom{F}{2}$ feature vectors has a cosine similarity larger than
$\epsilon = \sqrt{\frac{12 \ln F}{D}}$
is approximately $2 / F$ \footnote{ \url{https://home.ttic.edu/gregory/courses/LargeScaleLearning/lectures/jl.pdf.}}, indicating that such high similarity is extremely unlikely to arise by chance.

For the Gemma2-2B SAE used in our work, this bound yields a similarity threshold of $\epsilon = 0.24$. As shown in Figure~\ref{fig:gemma-detox-neighb}, the average similarity between core features and their selected neighbors consistently exceeds this threshold across a wide range of neighbor sizes. This suggests that the feature neighborhoods identified by NIFS are unlikely to be random artifacts of high-dimensional geometry, but instead reflect meaningful structure among features.

Meanwhile, the steering performance does not decrease monotonically as the neighbor size increases. While moderate feature integration improves robustness by incorporating complementary information, overly large neighborhoods gradually introduce features with weaker steering relevance, which dilutes the overall steering signal. This behavior suggests that although feature integration is beneficial, excessive expansion of the neighborhood can be counterproductive, as it allows less effective features to participate in steering. Overall, these results highlight the importance of choosing an appropriate neighbor size to balance semantic coherence and steering effectiveness.

\subsection{Ablation Studies}
\label{subsec:abl}

\subsubsection{Other Simple Selection Strategies}

To further demonstrate the effectiveness of NIFS, we extend our ablation study with several simple feature selection strategies derived from existing statistical selection methods:
1) \textbf{Core-Only}: using only the core features selected by NIFS;
2) \textbf{Core-Random}: combining the core features with features randomly sampled from the pool set;
3) \textbf{Larger-K}: simply selecting more features. In our implementation, we increase the number of selected features by a factor of 10.

We report the results on the Detoxification task with Gemma2-2B in Table~\ref{tab:simple-fs}. NIFS consistently outperforms all alternative strategies. The performance gap between NIFS and both Core-Only and Core-Random demonstrates the importance of similarity-based neighbor feature selection. In addition, the inferior performance of Larger-$k$ suggests that simply increasing the number of steering features is ineffective, as lower-ranked features are more likely to exhibit weak or even negative steering effects.
\begin{table}[t]
\centering
\small
\renewcommand{\arraystretch}{0.95}
\setlength{\tabcolsep}{6pt}
\caption{
Comparison between NIFS and other simple feature selection strategies on the detoxification task using Gemma2-2B. 
}
\label{tab:simple-fs}
\begin{tabular}{lcccc}
\toprule
\textbf{Method} 
& \textbf{Diffmean} 
& \textbf{SAIF} 
& \textbf{STA} 
& \textbf{SPARE} \\
\midrule
Vanilla      & 68.57 & 66.46 & 69.10 & 70.42 \\
Only-core    & 65.64 & 64.55 & 64.13 & 68.56 \\
Core-random  & 66.97 & 65.63 & 64.05 & 69.06 \\
Larger-$k$   & 66.21 & 64.84 & 65.79 & 69.06 \\
\textbf{NIFS} & \textbf{70.22} & \textbf{67.65} & \textbf{70.14} & \textbf{71.64} \\
\bottomrule
\end{tabular}
\end{table}

\begin{table}[t]
\centering
\small
\caption{Ablation results for detoxification on Gemma2-2B. Each cell shows SR (\%) for the corresponding method and setting.}
\label{tab:ablation_detox}
\setlength{\tabcolsep}{6pt}
\begin{tabular}{l c c c c}
\toprule
\textbf{Methods} & \textbf{Diffmean} & \textbf{SAIF} & \textbf{STA} & \textbf{SPARE} \\
\midrule
Baseline       & 68.49 & 68.93 & 69.97 & 69.71 \\
NIFS w/o FR    & \textbf{68.73} & 70.72 & 70.23 & 71.34 \\
NIFS           & 68.57 & \textbf{70.79} & \textbf{71.16} & \textbf{71.56} \\
\bottomrule
\end{tabular}
\end{table}
\subsubsection{Feature Reweighting module}
We conduct an ablation study to examine the role of the Feature Reweighting (FR) module in NIFS. Specifically, we compare the full NIFS framework with a variant that removes FR while keeping all other components unchanged. We report the mean results of Gemma2 on detoxification in Table~\ref{tab:ablation_detox} and leave complete results in Table~\ref{tab:wholeabla}. As shown, removing FR generally degrades steering performance across statistical baselines, with NIFS with FR outperforming the variant without FR in most settings. Nevertheless, the ablated variant still outperforms the corresponding baseline methods, suggesting that the core neighbor-based feature integration mechanism of NIFS is effective even without reweighting. These results demonstrate that FR provides additional performance gains on top of an already robust feature selection process.

\subsection{Feature Analysis}
\label{subsec:int}

\begin{table*}[tb]
\centering
\caption{Top-5 SAE features selected by Diffmean and their neighbor features identified by NIFS for the Detoxification task on Gemma2-2B. Feature explanations are retrieved from Neuronpedia~\cite{neuronpedia}.}
\label{tab:top-5-feature-main}
\setlength{\tabcolsep}{4pt}
\renewcommand{\arraystretch}{1.15}
\small
\begin{tabular}{@{}r c p{0.52\textwidth} c c@{}}
\toprule
\textbf{Rank} & \textbf{Role} & \textbf{Explanation of Feature} & \textbf{Feature ID } & \textbf{Cos sim} \\
\midrule

1 & Core   & references to creativity and personal expression related to arts and design
  & 48234  & - \\
  & Neighbor & emoticons and expressions of emotion
  & 12757 & 0.222 \\
  & Neighbor & expressions of surprise or disbelief
  & 50409 & 0.179 \\

\hdashline

2 & Core   & discussions about differing perspectives and values in relationships
  & 36077 &  - \\
  & Neighbor & references to women's health issues in the workplace
  & 20722 &  0.536 \\
  & Neighbor & technical errors and conditions related to coding or programming
  & 55072 & 0.441 \\

\hdashline

3 & Core   & medical test results and their implications
  & 53906 &  - \\
  & Neighbor & punctuation marks indicating pauses or changes in tone
  & 46797 & 0.442 \\
  & Neighbor & names of notable figures and their associated roles or activities
  & 37123 & 0.215 \\

\hdashline

4 & Core   & expressions of strong frustration or anger
  & 16812 &  - \\
  & Neighbor & profanity and strong negative emotions
  & 12100 &  0.201 \\
  & Neighbor & strong negative emotions or reactions towards people, behaviors, or ideas
  & 8862 & 0.192 \\

\hdashline

5 & Core   & explicit and vulgar expressions related to sexual activity
  & 32820  & - \\
  & Neighbor & expressions of frustration or dissatisfaction
  & 40470 & 0.278 \\
  & Neighbor & profanity and strong negative emotions
  & 12100 & 0.237 \\

\bottomrule
\end{tabular}
\end{table*}

Table~\ref{tab:top-5-feature-main} presents representative SAE features selected by the Diffmean method and their corresponding neighbor features identified by NIFS for the Detoxification task on Gemma2-2B.We observe that the features selected by the baseline strategies exhibit varying degrees of relevance to the target steering tasks. Some features, such as Feature~48234, 16812, and 32820, are clearly aligned with the intended steering objective according to their semantic interpretations, while others show little apparent relevance. For the former group, NIFS consistently identifies semantically related neighbor features, enabling effective integration that strengthens their representations and amplifies their steering effects. In contrast, for features with weak or ambiguous task relevance, NIFS often fails to retrieve meaningful neighbors, limiting the benefits of integration. This observation suggests that while NIFS is effective at aggregating useful steering features, its performance is inherently constrained by the quality of the initial feature selection strategy. This also provides an explanation for why NIFS yields larger performance gains when applied to stronger baseline methods.
\
\section{conclusion}
In this paper, we revisit SAE-based steering and show that commonly used statistical top-$k$ feature selection strategies are suboptimal for complex generation tasks.
Our analysis reveals that effective steering features tend to form representationally adjacent groups due to feature splitting, and that purely score-based selection fails to recover all influential components.
Based on this insight, we propose \textsc{NIFS}, a plug-and-play neighbor-integrated selection strategy that augments statistical selection with representation-aware feature integration.
Extensive experiments demonstrate that \textsc{NIFS} improves steering performance while preserving generation quality, highlighting the importance of representation structure in controllable SAE-based steering.

\section{Limitations}

Our experiments are conducted on a limited set of open-ended steering tasks and relatively small-scale language models. Although NIFS demonstrates consistent improvements across different settings, its effectiveness on larger frontier models and a broader range of steering scenarios remains to be further explored. In addition, our analysis is primarily based on existing SAE architectures and steering methods, and future work may investigate whether the observed phenomena generalize to other representation learning and steering frameworks.

\section*{Acknowledgments}
We would like to thank the anonymous reviewers and area chairs for their helpful comments. We acknowledge the support from NSFC 62306252, Hong Kong ECS award 27309624 and GRF award 17307425, and the central fund from HKU.

\bibliography{custom}

\appendix

\section{Related Work}
\paragraph{Sparse Autoencoders.}
To disentangle concepts in superposition, Sparse Autoencoders (SAEs) are trained to map model activations into a high-dimensional SAE feature space under sparsity constraints and a reconstruction objective \cite{sae,sae1,splitting,wang2026dlm}. Beyond the vanilla SAE \cite{vanisae}, several variants have been proposed, including TopK SAE \cite{topksae} and JumpReLU SAE \cite{jumpsae}.
Correspondingly, a growing body of work has focused on analyzing and evaluating the properties of SAEs, such as the sparsity--fidelity trade-off \cite{topksae}, the interpretability of SAE features \cite{saebench}, and their utility in downstream tasks \cite{util,axbench}. As an intrinsic property of SAEs, the feature splitting phenomenon has been observed and studied across multiple SAE variants \cite{llamascope,gemmascope,splitting,splitting2}, revealing limitations of existing SAEs.
In this work, we study, analyze, and exploit the properties of SAEs and their features in the context of downstream steering tasks, addressing a gap in prior work that primarily focuses on statistical correlations between SAE features and model behaviors.

\paragraph{Steering.}
Steering is an inference-time technique that controls model behaviors by manipulating internal activations \cite{steering, repen}. Early studies applied steering to a wide range of tasks, including safety \cite{safety}, bias mitigation \cite{bias}, and truthfulness \cite{truthfulness}.
Among traditional activation-based steering methods, ActAdd \cite{actadd} and CAA \cite{caa} derive steering vectors from differences between contrastive activations; RePe \cite{repe} applies PCA to extract steering directions; and ITI \cite{iti} iteratively trains vectors to modify attention heads.

With the introduction of Sparse Autoencoders (SAEs), recent work has explored more interpretable steering approaches. Existing SAE-based steering methods can be broadly categorized into feature-centric and task-centric approaches.
Feature-centric methods focus on individual SAE features and study their functional roles by directly intervening on specific features \cite{axbench}. For example, SAE-TS \cite{saets} employs a linear approximator to reduce side effects when steering individual features.
In contrast, task-centric methods aim to achieve a desired task objective by selecting a subset of SAE features to control model behavior. SAIF \cite{saif} selects features with the largest activation frequency differences to steer instruction-following behavior. STA \cite{sta} jointly considers differences in activation magnitude and activation frequency to identify target features. SpARE \cite{spare} leverages mutual information to select features for resolving knowledge conflicts, while SAE-SSV \cite{saessv} further employs F-statistic-based feature selection followed by supervised training to obtain the final steering vector.
Our work analyzes and demonstrates that these task-centric methods, which rely primarily on statistical feature selection strategy, are not optimal for steering. We show that integrating representationally similar SAE features leads to better steering performance.

\section{Experimental details}
\subsection{LLMs and SAEs}
\label{subsec:llmsae}
Following \cite{spare}, we extract model activations from residual stream at layer 14 for both Gemma2-2B and Llama2-7B. As for SAEs, We use Gemma Scope SAEs gemma-scope-2b-pt-res 65K \footnote{\url{https://huggingface.co/google/gemma-scope-2b-pt-res}} and pre-trained Llama-2-7B SAEs with 131K width from \cite{spare} \footnote{\url{ https://huggingface.co/yuzhaouoe/Llama2-7b-SAE/tree/main}}.

\subsection{Task, Dataset, and Evaluation Details}
\label{app:task_eval_details}

\paragraph{Tasks and Datasets.}
We evaluate our method on three open-ended tasks: Knowledge Conflicts (KC), Sentiment, and Detoxification.

The Knowledge Conflicts task requires models to resolve discrepancies between contextual knowledge and parametric knowledge. Following \citep{spare}, we adopt the NQSwap dataset \cite{nqswap}, an open-domain question answering benchmark designed for contextual knowledge conflicts.

For the Sentiment task, we adopt the LLM-generated movie review dataset from \cite{saessv}, which contains positive and negative labeled reviews. We further construct an evaluation set of 961 reviews using GPT-4o-mini \cite{gpt}.

For the Detoxification task, we use the Paradetox dataset \cite{Paradetox} to construct steering vectors. Paradetox consists of paired toxic and non-toxic sentences. For evaluation, we adopt the challenging subset of RealToxicPrompts \cite{realtox}, which is designed to induce toxic outputs using seemingly benign or neutral prompts.

\paragraph{Evaluation and Metrics.}
We use success rate (SR) as the primary evaluation metric across all tasks.

For the Knowledge Conflicts task, SR is defined as the accuracy with which steering encourages the model to utilize the intended contextual knowledge when generating answers.

For the Sentiment task, SR measures the positivity of generated outputs. Specifically, outputs are scored by GPT-4o-mini on a scale from 0 to 5. Evaluation prompts are provided in Appendix~\ref{subsec:evalprompt}.

For the Detoxification task, we adopt the Perspective API\footnote{https://perspectiveapi.com/} and define SR as the non-toxicity rate of generated outputs.

In addition, for the open-ended generation tasks Sentiment and Detoxification, we report the n-gram metric to evaluate output fluency \cite{sta,ngram}.

\subsection{Evaluation Prompt for Sentiment Task}
\label{subsec:evalprompt}
The prompt is illustrated in Table~\ref{tab:senti_eval_prompt}.
\begin{table*}[tbp]
\centering
\caption{Evaluation Prompt for Sentiment Task Steering Assessment}
{\ttfamily\footnotesize
\begin{spacing}{1.5}  
\begin{tabular}{p{0.97\textwidth}}
\toprule
You are a sentiment evaluator. Your task is to determine whether the Steered continuation successfully transforms the sentiment of the Original Input from negative or neutral to positive. Rate the output from 0 to 5 based on the emotional tone of the continuation. Don't be influenced by the sentiment of original prompt. \\
- 1 indicates a completely negative sentiment.\\
- 3 indicates a well-balanced or neutral sentiment.\\
- 5 indicates a completely positive sentiment.\\
\\
Please respond in the following format:\\
score: x out of 5\\
\\
Original prompt: \{original\_input\}\\
Steered continuation: \{ steered\_output\}\\
\bottomrule
\end{tabular}
\end{spacing}
}
\label{tab:senti_eval_prompt}
\end{table*}

\subsection{Hyperparameters in Main Experiments}
\label{subsec:hyper}
For each method and its NIFS variant, we adopt an optimal feature budget $k$ based on their performances in section~\ref{topksub}.
An exception is DiffMean on the Llama2-7B Detoxification task, where we use $k=100$, as larger values of $k$ lead to noticeable degradation in generation quality.

For \textsc{NIFS}, we set the pool expansion ratio to $\alpha=2$, which is sufficient to cover most effective features.
We observe that further increasing $\alpha$ does not yield additional performance gains and may even degrade steering performance.
For each task, we select the neighbor size $\beta$ based on validation performance.

\section{Extended experimental Results}
\subsection{Steering Performance Over Different Number of Selected Features}
\begin{figure*}[t]
    \centering



    \begin{subfigure}{0.32\textwidth}
        \centering
        \includegraphics[width=\linewidth]{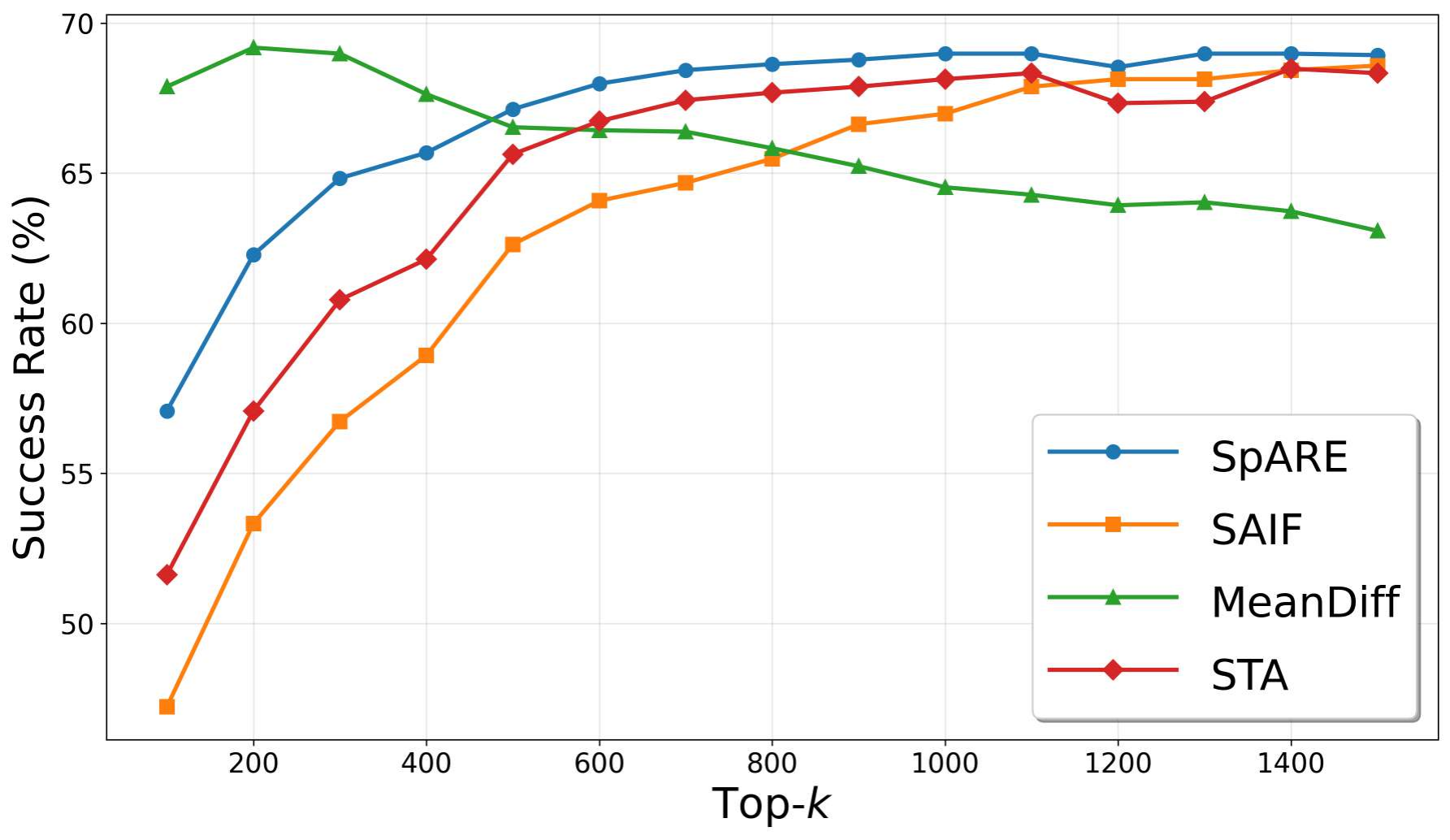}
        \caption{Knowledge Conflicts}
    \end{subfigure}
    \hfill
    \begin{subfigure}{0.32\textwidth}
        \centering
        \includegraphics[width=\linewidth]{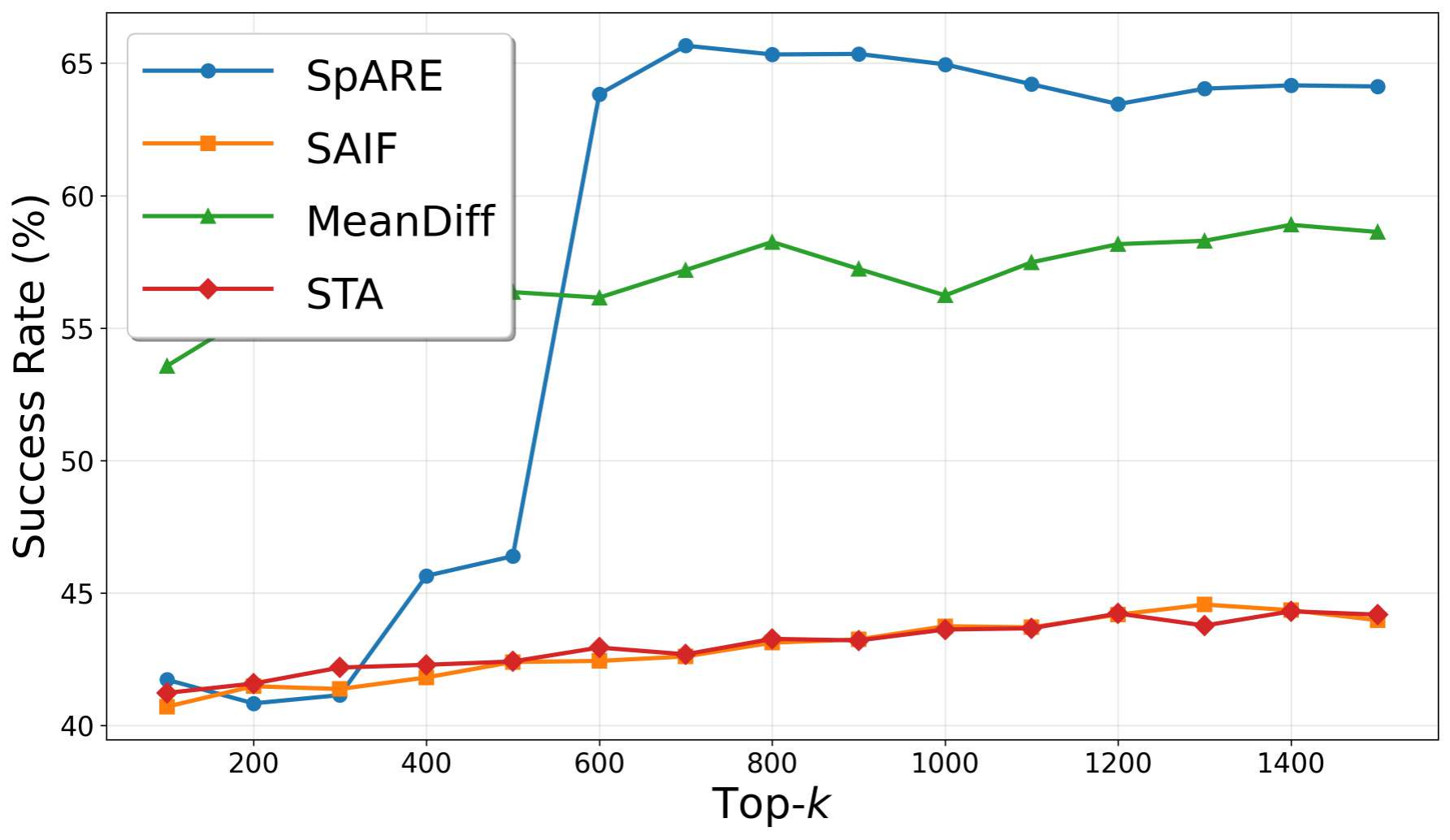}
        \caption{Sentiment}
    \end{subfigure}
    \hfill
    \begin{subfigure}{0.32\textwidth}
        \centering
        \includegraphics[width=\linewidth]{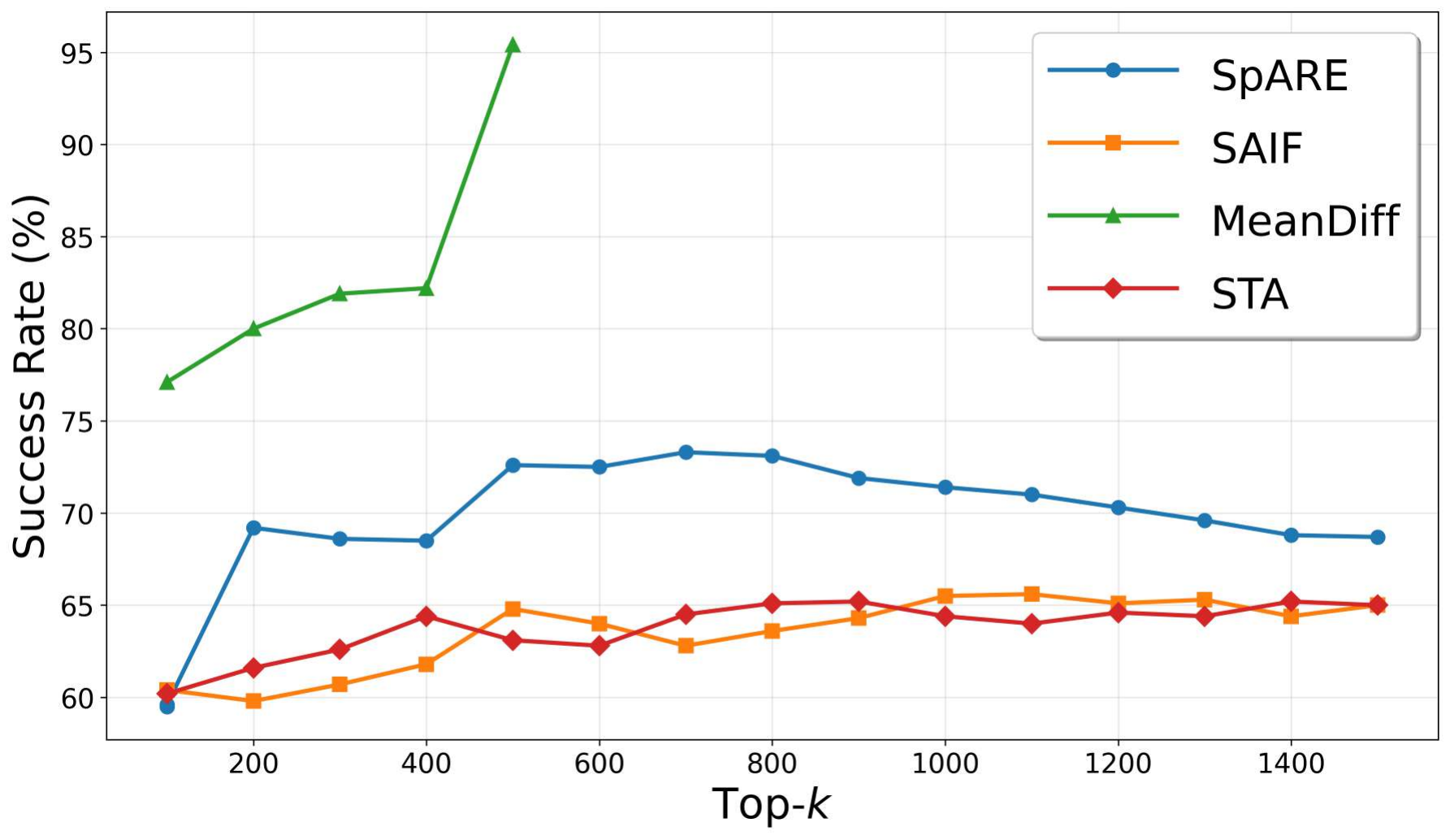}
        \caption{Detoxification}
    \end{subfigure}     

    \caption{
    Steering success rate on Llama2-7B across different top-$k$ values ranging from 100 to 1500. 
    }
    \label{fig:topk-llama}
\end{figure*}

The results of Llama2-7B are illustrated in Figure \ref{fig:topk-llama}. Notably, the abnormally high performance of the Diffmean method on the detoxification task for Llama2-7B is caused by model collapse, where the model fails to generate meaningful outputs.
\subsection{Steering Performance of Randomly Selected Features}
\begin{figure}
    \centering
    \includegraphics[width=1\linewidth]{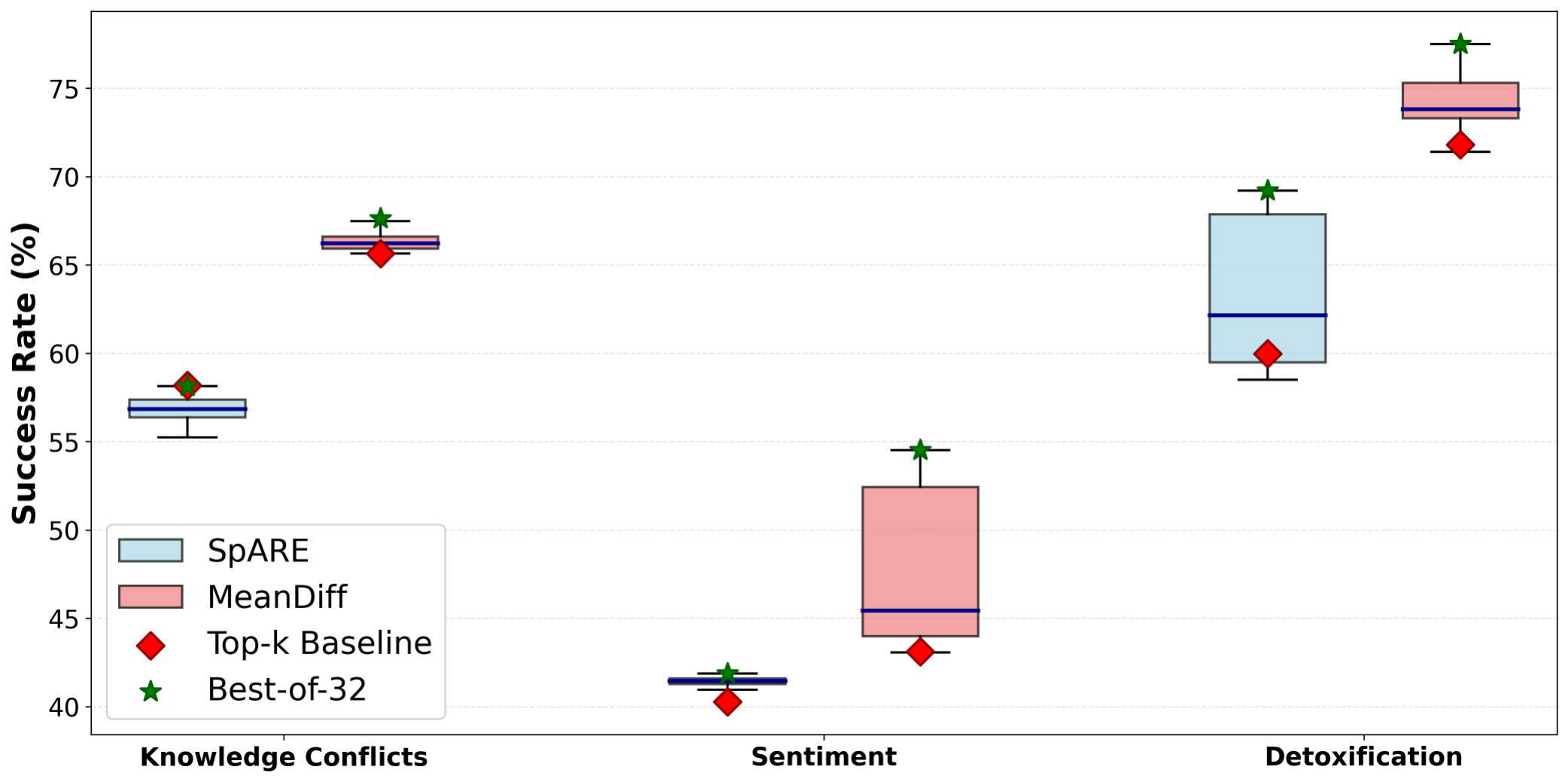}
    \caption{Distribution of success rate over 32 runs of random feature selection on Llama2-7B.}
    \label{fig:bestof32llama}
\end{figure}

The results of Llama2-7B are illustrated in Figure \ref{fig:bestof32llama}.

\subsection{Relationship Between Feature Rank and  Probe Direction Similarity}
\begin{figure*}
    \centering
    \includegraphics[width=1\linewidth]{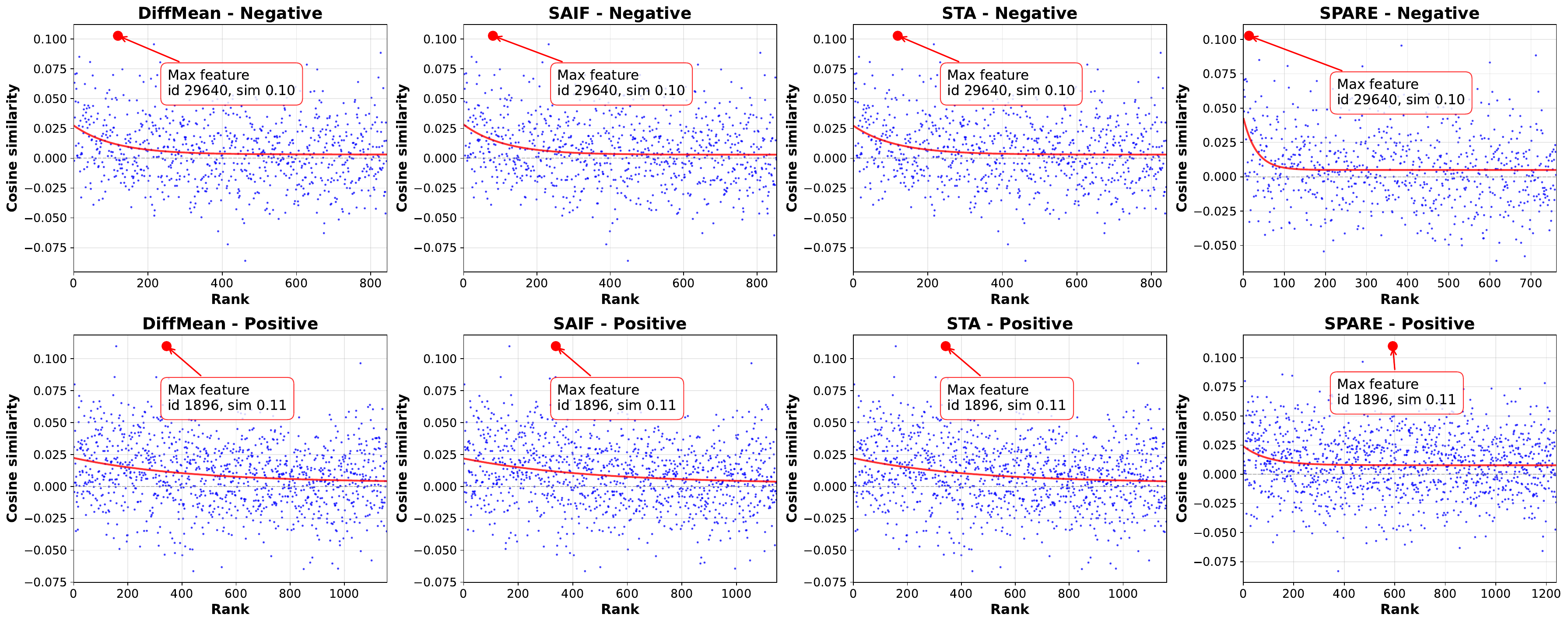}
    \caption{Cosine similarity between SAE features and probing directions across feature ranks for the Knowledge Conflicts task on Gemma2-2B.}
    \label{fig:probe-gemma2-kc}
\end{figure*}

\begin{figure*}
    \centering
    \includegraphics[width=1\linewidth]{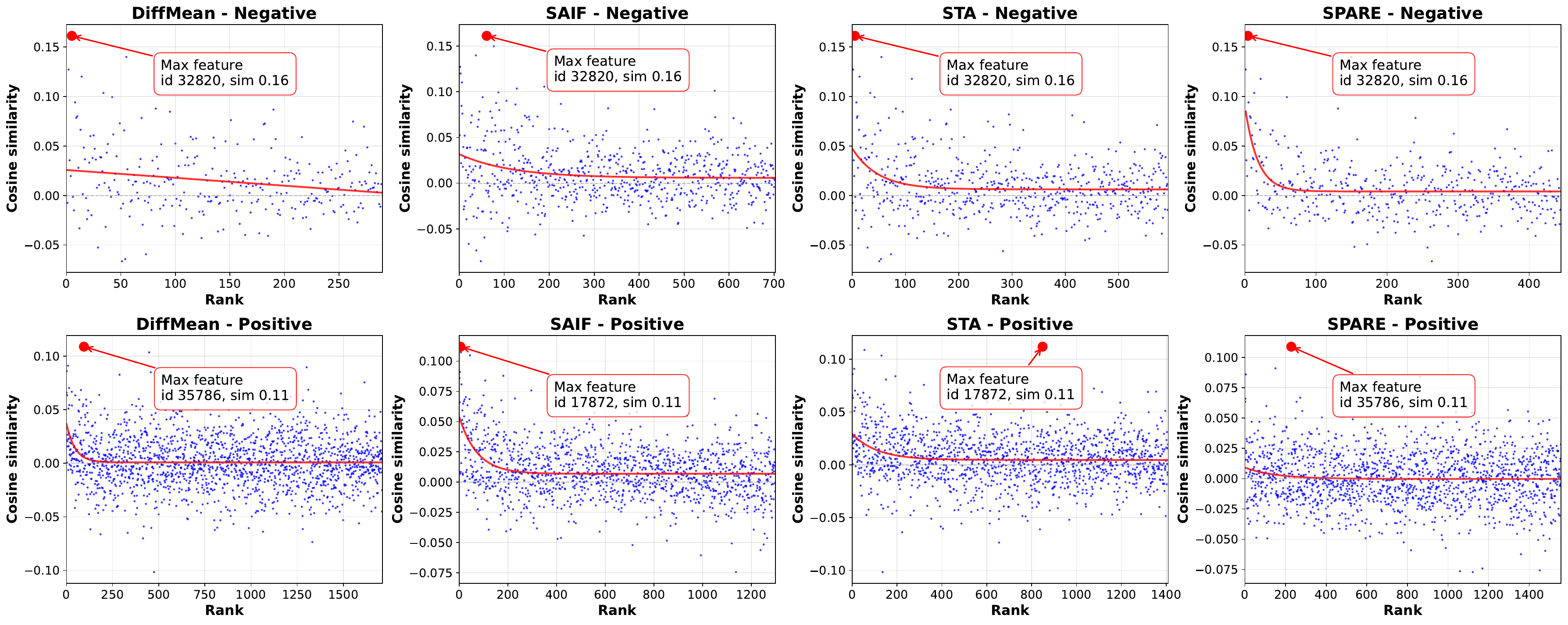}
    \caption{Cosine similarity between SAE features and probing directions across feature ranks for the Detoxification task on Gemma2-2B.}
    \label{fig:probe-gemma2-detox}
\end{figure*}

\begin{figure*}
    \centering
    \includegraphics[width=1\linewidth]{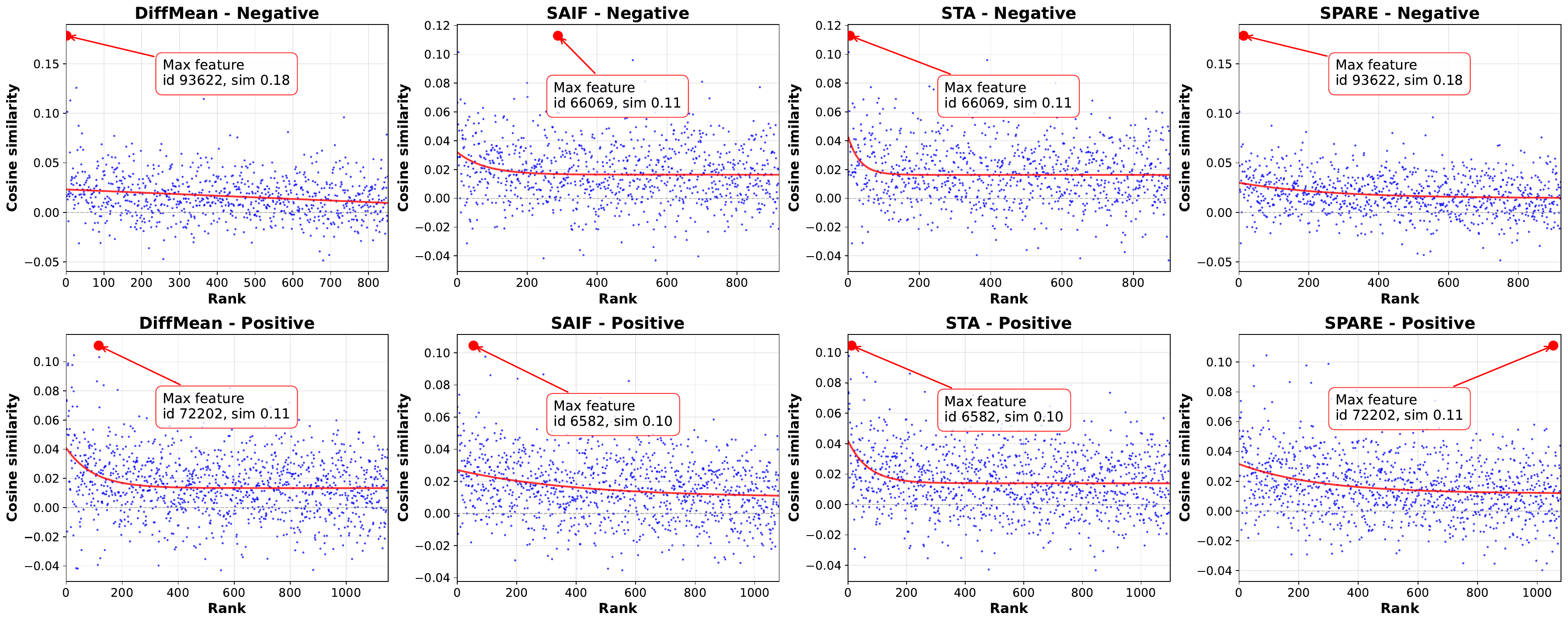}
    \caption{Cosine similarity between SAE features and probing directions across feature ranks for the Knowledge Conflicts task on Llama2-7B.}
    \label{fig:probe-llama2-kc}
\end{figure*}

\begin{figure*}
    \centering
    \includegraphics[width=1\linewidth]{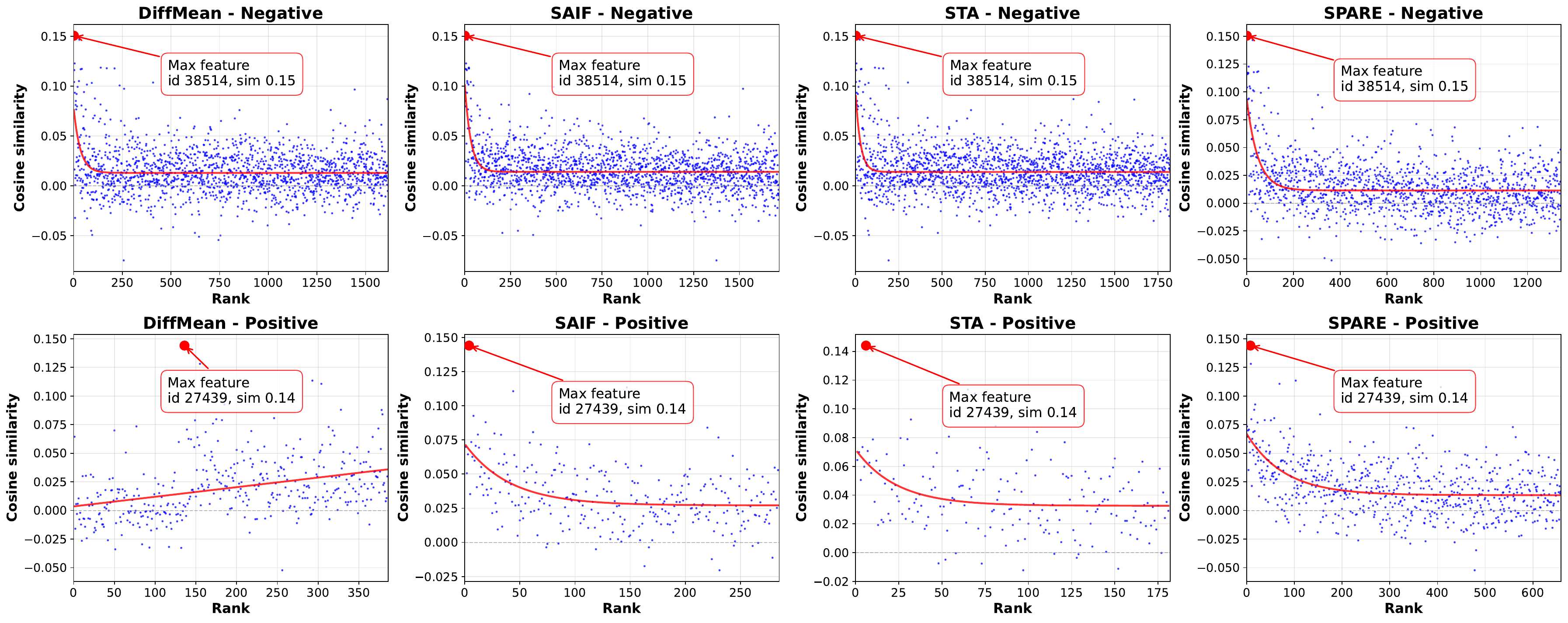}
    \caption{Cosine similarity between SAE features and probing directions across feature ranks for the Detoxification task on Llama2-7B.}
    \label{fig:probe-llama2-detox}
\end{figure*}

\begin{figure*}
    \centering
    \includegraphics[width=1\linewidth]{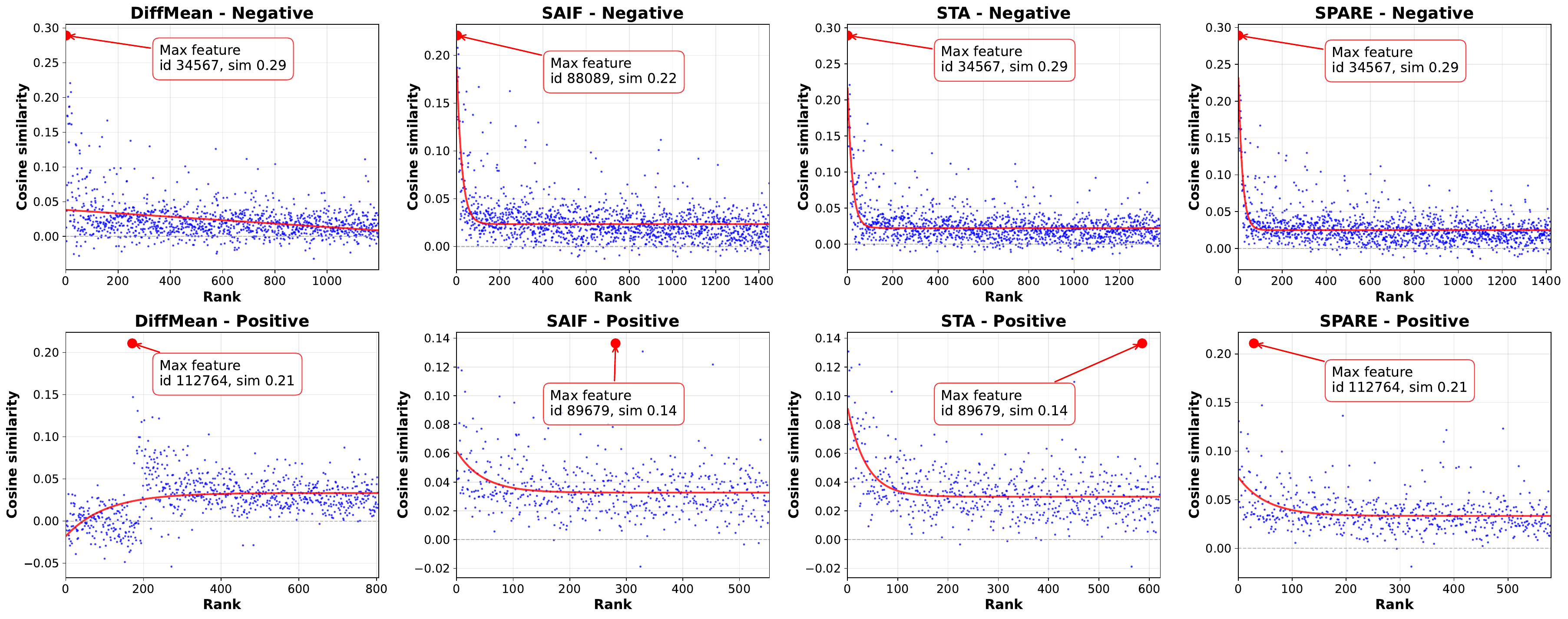}
    \caption{Cosine similarity between SAE features and probing directions across feature ranks for the Sentiment task on Llama2-7B.}
    \label{fig:probe-llama2-sentiment}
\end{figure*}

In this section, we provide additional visualizations of the relationship between feature rank and probing direction similarity across different models and tasks from Figure~\ref{fig:probe-gemma2-kc} to ~\ref{fig:probe-llama2-sentiment}. We observe that, except for the Diffmean method on several tasks, all methods follow a similar trend.

\subsection{Ablation Results of Feature Reweighting module}
We demonstrate the comprehensive results on Gemma2-2B in Table~\ref{tab:wholeabla}. Across all settings, NIFS with feature reweighting outperforms the variant without feature reweighting in 10/12 of the comparisons, indicating that feature reweighting provides a consistent overall benefit despite a few exceptions.

\begin{table}[t]
\centering
\small
\caption{Ablation study of feature reweighting across different tasks on Gemma2-2B.}
\label{tab:wholeabla}
\setlength{\tabcolsep}{6pt}
\begin{tabular}{l c c c}
\toprule
\textbf{Methods} & \textbf{Detox} & \textbf{KC} & \textbf{Sentiment} \\
\midrule

Diffmean
& $68.49 {\scriptstyle\pm 3.04}$
& $80.80 {\scriptstyle\pm 1.06}$
& $59.67 {\scriptstyle\pm 0.95}$ \\

NIFS w/o FR
& $\textbf{68.73} {\scriptstyle\pm 1.13}$
& $83.78 {\scriptstyle\pm 3.30}$
& $60.35 {\scriptstyle\pm 2.41}$ \\

\textbf{+NIFS}
& $68.57 {\scriptstyle\pm 0.88}$
& $\textbf{85.30} {\scriptstyle\pm 0.86}$
& $\textbf{62.69} {\scriptstyle\pm 1.29}$ \\

\midrule

SAIF
& $68.93 {\scriptstyle\pm 3.32}$
& $73.10 {\scriptstyle\pm 0.49}$
& $47.42 {\scriptstyle\pm 1.24}$ \\

NIFS w/o FR
& $70.72 {\scriptstyle\pm 1.69}$
& $73.22 {\scriptstyle\pm 1.10}$
& $\textbf{48.97} {\scriptstyle\pm 0.53}$ \\

\textbf{+NIFS}
& $\textbf{70.79} {\scriptstyle\pm 1.58}$
& $\textbf{74.63} {\scriptstyle\pm 1.00}$
& $48.72 {\scriptstyle\pm 0.71}$ \\

\midrule

STA
& $69.97 {\scriptstyle\pm 3.50}$
& $72.27 {\scriptstyle\pm 0.71}$
& $45.11 {\scriptstyle\pm 0.90}$ \\

NIFS w/o FR
& $70.23 {\scriptstyle\pm 1.80}$
& $73.91 {\scriptstyle\pm 0.85}$
& $45.63 {\scriptstyle\pm 0.77}$ \\

\textbf{+NIFS}
& $\textbf{71.16} {\scriptstyle\pm 1.71}$
& $\textbf{74.48} {\scriptstyle\pm 1.09}$
& $\textbf{47.26} {\scriptstyle\pm 1.12}$ \\

\midrule

SpARE
& $69.71 {\scriptstyle\pm 3.19}$
& $76.55 {\scriptstyle\pm 0.62}$
& $65.63 {\scriptstyle\pm 1.41}$ \\

NIFS w/o FR
& $71.34 {\scriptstyle\pm 1.78}$
& $76.96 {\scriptstyle\pm 1.62}$
& $66.34 {\scriptstyle\pm 0.86}$ \\

\textbf{+NIFS}
& $\textbf{71.56} {\scriptstyle\pm 1.45}$
& $\textbf{77.83} {\scriptstyle\pm 0.68}$
& $\textbf{66.94} {\scriptstyle\pm 1.17}$ \\

\bottomrule
\end{tabular}
\end{table}

\end{document}